\documentclass[table,10pt,twocolumn,letterpaper]{article} 

\usepackage{cvpr}              

\usepackage{url}

\usepackage{booktabs}
\usepackage{chngpage}
\usepackage{multirow}
\usepackage{tabularx}
\usepackage{makecell}
\usepackage{comment}
\usepackage{rotating}
\usepackage{caption}
\usepackage{subcaption}
\usepackage[dvipsnames]{xcolor}
\usepackage{etoolbox}
\usepackage{listings}
\usepackage[cachedir=.,frozencache]{minted}
\usepackage[accsupp]{axessibility} 
\usepackage[normalem]{ulem}
\usepackage[detect-all,separate-uncertainty = true]{siunitx}
\robustify{\bfseries}
\robustify{\uline}
\makeatletter
\robustify\@latex@warning@no@line
\makeatother
\usepackage{authblk}  

\makeatletter
\renewcommand\AB@affilsepx{, \protect\Affilfont}
\makeatother

\definecolor{cvprblue}{rgb}{0.21,0.49,0.74}
\usepackage[pagebackref,breaklinks,colorlinks,citecolor=cvprblue]{hyperref}
\usepackage[capitalize]{cleveref}

\def\paperID{15990} 
\def\confName{CVPR}
\def\confYear{2024}

\crefformat{section}{\S#2#1#3}

\title{\modelname{}: A Vision Foundation Model for the Tree of Life}

\author[1*$\dagger$]{Samuel~Stevens}
\author[1*]{Jiaman~Wu}
\author[1]{Matthew~J~Thompson}
\author[1]{Elizabeth~G~Campolongo}
\author[1]{Chan~Hee~Song}
\author[1]{David~Edward~Carlyn}
\author[2]{Li~Dong}
\author[3]{Wasila~M~Dahdul}
\author[4]{Charles~Stewart}
\author[1]{Tanya~Berger-Wolf}
\author[1]{Wei-Lun~Chao}
\author[1$\dagger$]{Yu~Su}
\affil[1]{The~Ohio~State~University}
\affil[2]{Microsoft~Research}
\affil[3]{University~of~California,~Irvine}
\affil[4]{Rensselaer~Polytechnic~Institute\protect\\}

\newcommand{\modelname}{\textsc{BioCLIP}}
\newcommand{\datasetname}{\textsc{TreeOfLife-10M}}
\newcommand{\datasetsmall}{\textsc{TreeOfLife-1M}}
\newcommand{\datasetshortname}{\textsc{ToL-10M}}
\newcommand{\datasetshortsmall}{\textsc{ToL-1M}}
\newcommand{\bioscan}{\textsc{Bioscan}-1M}
\newcommand{\rarespecies}{\textsc{Rare Species}}
\newcommand{\vertical}[1]{\multicolumn{1}{c}{\rotatebox{90}{#1}}}

\newcommand{\sithead}[1]{\text{\textbf{#1}}}

\newcommand{\second}[1]{\textit{#1}}

\newcommand\extrafootertext[2]{%
    \bgroup
    \renewcommand\thefootnote{\fnsymbol{footnote}}%
    \renewcommand\thempfootnote{\fnsymbol{mpfootnote}}%
    \footnotetext[#1]{#2}%
    \egroup
}

\begin{document}

\pagenumbering{gobble}

\maketitle

\extrafootertext{1}{Equal contribution. $^{\dagger}$\texttt{\small \{stevens.994,su.809\}@osu.edu}}


\begin{abstract}
Images of the natural world, collected by a variety of cameras, from drones to individual phones, are increasingly abundant sources of biological information. 
There is an explosion of computational methods and tools, particularly computer vision, for extracting biologically relevant information from images for science and conservation. 
Yet most of these are bespoke approaches designed for a specific task and are not easily adaptable or extendable to new questions, contexts, and datasets. 
\emph{A vision model for general organismal biology questions on images is of timely need.}
To approach this, we curate and release \datasetname{}, the largest and most diverse ML-ready dataset of biology images. 
We then develop \modelname{}, a foundation model for the tree of life, leveraging the unique properties of biology captured by \datasetname{}, namely the abundance and variety of images of plants, animals, and fungi, together with the availability of rich structured biological knowledge.
We rigorously benchmark our approach on diverse fine-grained biology classification tasks and find that \modelname{} consistently and substantially outperforms existing baselines (by 16\% to 17\% absolute). 
Intrinsic evaluation reveals that \modelname{} has learned a hierarchical representation conforming to the tree of life, shedding light on its strong generalizability.\footnote{\href{https://imageomics.github.io/bioclip/}{\texttt{imageomics.github.io/bioclip}} has models, data and code.}
\end{abstract}

\vspace{-12pt}


\begin{figure*}[t]
    \centering
    \small
    \begin{subfigure}[b]{0.7\textwidth}
        \includegraphics[width=\textwidth]{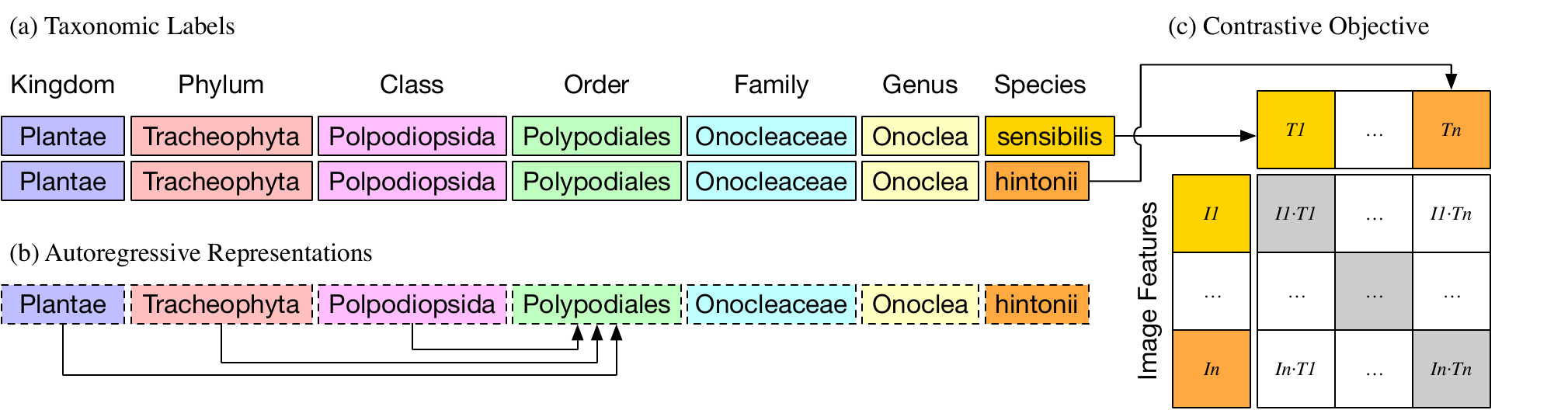}
    \end{subfigure}
    \hfill
    \begin{subfigure}[b]{0.14\textwidth}
        \addtocounter{subfigure}{3}
        \includegraphics[width=\textwidth]{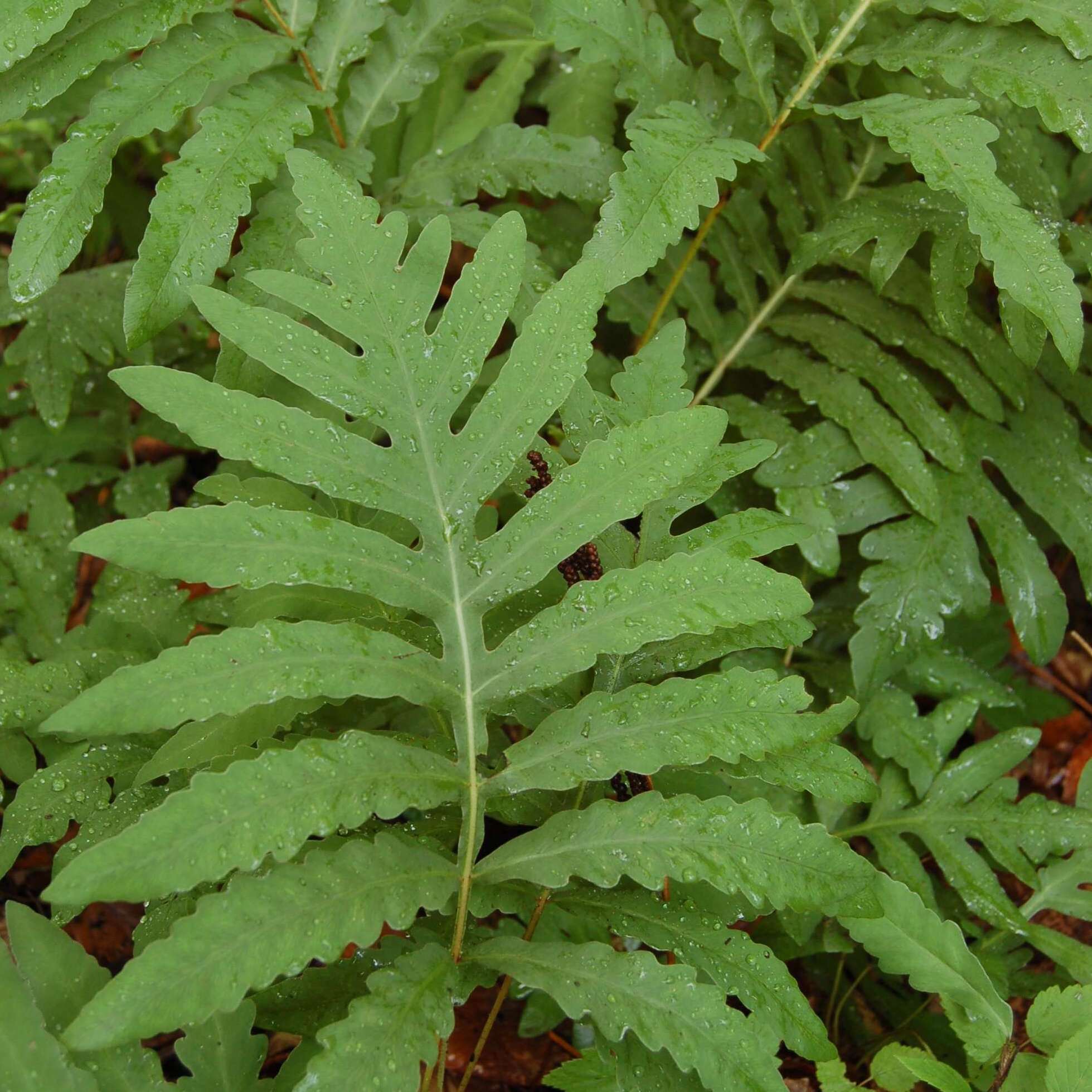}
        \caption{\textit{Onoclea sensibilis}}
        \label{subfig:sensibilis}
    \end{subfigure}
    \hfill
    \begin{subfigure}[b]{0.14\textwidth}
        \includegraphics[width=\textwidth]{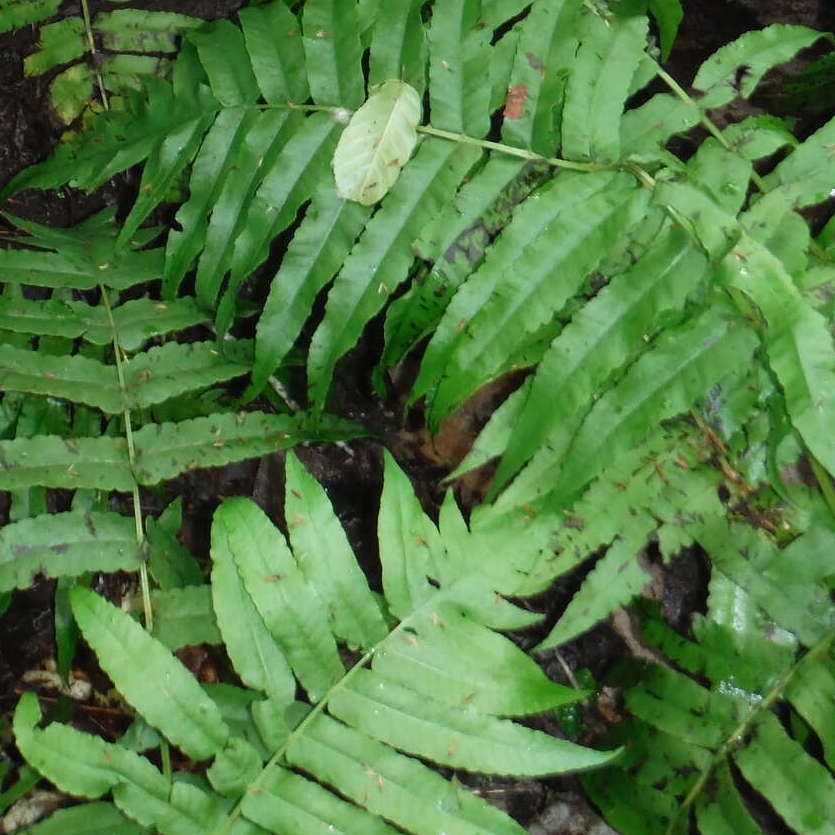}
        \caption{\textit{Onoclea hintonii}}
        \label{subfig:hintonii}
    \end{subfigure}
    \vskip -6pt
    \caption{
    (a) Two taxa, or taxonomic labels, for two different plants, \textit{Onoclea sensibilis} (d) and \textit{Onoclea hintonii} (e). 
    These taxa are identical except for the species. 
    (b) The autoregressive text encoder naturally encodes the hierarchical structure of the taxonomy. 
    See how the Order token(s) (Polypodiales) can incorporate information from the Kingdom, Phylum and Class tokens, but nothing later in the hierarchy. 
    This helps align the visual representations to this same hierarchical structure (see \cref{subsec:intrinsic-eval}).
    (c) These hierarchical representations of taxonomic labels are fed into the standard contrastive pre-training objective and are matched with image representations (d) and (e).
    }
    \label{fig:hook}
    \vskip -8pt
\end{figure*}

\section{Introduction}

Digital images and computer vision are quickly becoming pervasively used tools to study the natural world, from evolutionary biology~\citep{borowiec2022deep, lurig2021computer} to ecology and biodiversity~\citep{tuia2022perspectives,beery_scaling_2021,steenweg_scaling-up_2017}.
The capability to rapidly convert vast quantities of images from museums~\citep{pearson2020machine}, camera traps~\citep{beery2020iwildcam,beery2021iwildcam,steenweg_scaling-up_2017,norouzzadeh_deep_2021,ahumada_wildlife_2020}, and citizen science platforms~\citep{hoye2021deep, nugent2018inaturalist, sullivan2014ebird, antonelli_integrating_2023, mckinley_citizen_2017, sullivan_ebird_2014, swanson_snapshot_2015, parham_animal_2017, simpson_zooniverse_2014, van2015building,inat2017,norman_undersea_2017} into actionable information (e.g., species classification, individual identification, and trait detection) has accelerated and enabled new advances in tasks such as species delineation~\citep{hansen2020}, understanding mechanisms of adaptation~\citep{hoyal2019deep, ezray2019}, abundance and population structure estimation~\citep{hoye2021deep, teng2023bird,norman_undersea_2017,araujo_improving_2022}, and biodiversity monitoring and conservation~\citep{tuia2022perspectives}.  

However, applying computer vision to answer a biological question is still a laborious task requiring substantial machine learning expertise and effort---biologists must manually label sufficient data for the specific taxa and task of interest, and find and train a suitable model for the task. 
Meanwhile, foundation models \citep{bommasani2021opportunities} such as CLIP  \citep{radford2021learning} and GPT-3 \citep{brown2020language} are extraordinarily valuable by enabling zero-shot or few-shot learning for a wide range of tasks.
An analogous vision foundation model for biology should be useful for tasks spanning the \textit{entire} tree of life \citep{hinchliff2015synthesis, ToL_2007} instead of just the taxa it was trained on. 
Such a model would significantly lower the barrier to apply AI to biology.

In this work, we aim to develop such a vision foundation model for the tree of life. 
To be broadly useful for real-world biology tasks, this model should meet the following criteria. 
First, it should \textbf{generalize to the entire tree of life}, where possible, to ensure it supports researchers studying many different clades rather than a niche. 
Furthermore, it is infeasible to collect training data that covers the millions of known taxa~\citep{hobern2021towards, iucn2022}, so the model must generalize to taxa not present in training data. 
Second, it should learn \textbf{fine-grained representations} of images of organisms as biology frequently engages with organisms that are visually similar, like closely related species within the same genus \citep{pinho2022} or species mimicking others' appearances for a fitness advantage \citep{hoyal2019deep}. 
This fine-grained granularity is crucial because the tree of life organizes living things into both broad categories (animal, fungus, and plant) and very fine-grained ones (see \cref{fig:hook}).
Finally, due to the high cost of data collection and labeling in biology, \textbf{strong performance in the low-data regime} (i.e., zero-shot or few-shot) is critical. 

While the goals of \textbf{generalization}, \textbf{fine-grained classification}, and \textbf{data efficiency} are not new in computer vision, existing general-domain vision models~\citep{radford2021learning, yuan2021florence, oquab2023dinov2} trained on hundreds of millions of images fall short when applied to evolutionary biology and ecology. 
Specifically, existing vision models produce \textit{general} fine-grained representations, useful for comparing common organisms like dogs and wolves, but not for more fine-grained comparisons, e.g., \textit{Onoclea sensibilis} and \textit{Onoclea hintonii} (see \cref{fig:hook}).

We identify two major barriers to developing a vision foundation model for biology. First, there is a need for suitable \textbf{pre-training datasets}: existing datasets \cite{wah2011cub, inat2017, inat2021, gharaee2023step} lack either scale, diversity, or fine-grained labels. Second, there is a need to investigate suitable \textbf{pre-training strategies} that leverage special properties of the biology domain to better achieve the three pivotal goals, e.g., the tree of life taxonomy, which is insufficiently considered in mainstream pre-training algorithms \cite{liu2021swin,oquab2023dinov2,radford2021learning}.


In light of these goals and challenges in achieving them, we introduce 1) \textbf{\datasetname{}}, a large-scale ML-ready biology image dataset, and 2) \textbf{\modelname}, a vision foundation model for the tree of life, trained with suitable use of taxa in \datasetname. We outline the contributions, conceptual framework, and design decisions below:

\noindent \textbf{\datasetname{}: a large-scale, diverse ML-ready biology image dataset}. 
We curate and release the largest ML-ready dataset to-date of biology images with associated taxonomic labels, containing over \num{10} million images covering \num{454} thousand taxa in the tree of life.\footnote{
    By ML-ready, we mean the data is standardized in a format suitable for training ML models and is readily available for downloading.
}
In comparison, the current largest ML-ready biology image dataset, iNat21 \citep{inat2021}, contains only \num{2.7} million images covering \num{10} thousand taxa. 
\datasetname{} integrates existing high-quality datasets like iNat21 and \bioscan{} \citep{gharaee2023step}.
More importantly, it includes newly curated images from the Encyclopedia of Life (\href{https://eol.org}{eol.org}), which supplies most of \datasetname{'s} data diversity.
Every image in \datasetname{} is labeled with its taxonomic hierarchy to the finest level possible, as well as higher taxonomic ranks in the tree of life (see \cref{fig:hook,tab:classname} for examples of taxonomic ranks and labels).
\datasetname{} enables training \modelname{} and future biology foundation models.

\noindent \textbf{\modelname{}: a vision foundation model for the tree of life}. 
With a large-scale labeled dataset like \datasetname{,} a standard, intuitive training strategy (as adopted by other vision models like ResNet50 \citep{he2016deep} and Swin Transformer \citep{liu2021swin}) is to use a supervised classification objective and learn to predict the taxonomic indices from an image. 
However, this fails to recognize and leverage the rich structure of taxonomic labels---taxa do not exist in isolation but are interconnected in a comprehensive taxonomy. 
Consequently, a model trained via plain supervised classification may not generalize well to taxa unseen in training, nor could it support zero-shot classification of unseen taxa.

Instead, we propose a novel strategy \textit{combining CLIP-style multimodal contrastive learning~\cite{radford2021learning} with the rich biological taxonomy} for \modelname{}. We ``flatten'' the taxonomy from Kingdom to the distal-most taxon rank into a string called \textit{taxonomic name}, and use the CLIP contrastive learning objective to learn to match images with their corresponding taxonomic names. 
Intuitively, this helps the model generalize to unseen taxa---even if the model has not seen a species, it has likely learned a reasonable representation for that species' genus or family (see \cref{fig:hook}).
\modelname{} also supports zero-shot classification with taxonomic names of unseen taxa. 
We further propose, and demonstrate the effectiveness of, a \textit{mixed text type} training strategy; by mixing different text types (e.g., taxonomic vs.\ scientific vs.\ common names) during training, we retain the generalization from taxonomic names while being more flexibility at test time.
For example, \modelname{} still excels even if only common species names are offered by downstream users.

\noindent \textbf{Comprehensive benchmarking}.
We comprehensively evaluate \modelname{} on \num{10} fine-grained image classification datasets covering animals, plants, and fungi, including a newly curated \rarespecies{} dataset unseen in training. 
\modelname{} achieves strong performance in both zero-shot and few-shot settings and substantially outperforms both CLIP~\citep{radford2021learning} and OpenCLIP~\citep{ilharco2021openclip}, leading to an average absolute improvement of \textbf{17\%} (zero-shot) and \textbf{16\%} (few-shot).
Intrinsic analysis further reveals that \modelname{} has learned a more fine-grained hierarchical representation conforming to the tree of life, explaining its superior generalization.

\begin{table*}[ht]
    \centering
    \small
    \setlength\tabcolsep{3pt}
    \scalebox{0.85}{
    \begin{tabularx}{\textwidth}{lXrS[table-format=6.0]}
        \toprule
        \textbf{Dataset} & \textbf{Description} & \textbf{Images} & \textbf{Unique Classes} \\
        \midrule
        iNat21 & Citizen scientist labeled image dataset from \href{https://inaturalist.org}{iNaturalist} for fine-grained classification. & 2.7M & 10000  \\
        \midrule
        \bioscan{} & Expert labeled image dataset of insects for classification. & 1.1M & 7831 \\
        \midrule
        EOL & A new dataset with citizen scientist images sourced from \href{https://eol.org}{Encyclopedia of Life} and taxonomic labels standardized by us. & 6.6M & 448910 \\
        \midrule
        \textbf{\datasetname{}} & \textbf{Largest-to-date ML-ready dataset of biology images with taxonomic labels.} & \textbf{10.4M} & \bfseries 454103 \\
        \bottomrule
    \end{tabularx}
    }
    \vskip -6pt
    \caption{
        Training data sources used in \datasetname{.}
        We integrate and canonicalize taxonomic labels across the sources (\cref{subsec:metadata}).
    }
    \label{tab:training-data}
\end{table*}

\section{\datasetname{}}

Recent work has shown that data quality and diversity are critical when training CLIP models \citep{fang2022data, nguyen2022quality,gadre2023datacomp}.
We curate \datasetname{,} the most diverse large-scale public ML-ready dataset for computer vision models in biology. 


\subsection{Images}




The largest ML-ready biology image dataset is iNat21 \citep{inat2021} with \num{2.7}M images of \num{10}K species.
Despite this class breadth compared to popular general-domain datasets like ImageNet-1K \citep{russakovsky2015imagenet}, \num{10}K species is limited for biology. 
The International Union for Conservation of Nature (IUCN) reported over \num{2}M total described species in 2022, with over \num{10}K bird species and over \num{10}K reptile species alone \citep{iucn2022}.
iNat21's species diversity limits its potential for training a foundation model for the entire tree of life.


Motivated to find high-quality biology images with a focus on species diversity, we turn to the Encyclopedia of Life project (EOL; \href{https://eol.org}{eol.org}).
EOL collaborates with a variety of institutions to gather and label millions of images.
We download \num{6.6}M images from EOL and expand our dataset to cover an additional $\mathbf{440}$K taxa. 
Species are not evenly distributed among the different subtrees in the tree of life; insects (of the class \textit{Insecta} with \num{1}M+ species), birds (of the class \textit{Aves} with \num{10}K+ species) and reptiles (of the class \textit{Reptilia} with \num{10}K+ species) are examples of highly diverse subtrees with many more species.
To help a foundation model learn extremely fine-grained visual representations for insects, we also incorporate \bioscan{} \citep{gharaee2023step}, a recent dataset of \num{1}M lab images of insects, covering \num{494} different families.\footnote{
We note that \bioscan{}'s label granularity may still be limited for insects. 
\num{98.6}\% of \bioscan{'s} images are labeled to the family level but only \num{22.5}\% and \num{7.5}\% of the images have genus or species indicated, respectively. 
Lack of label granularity is an inherent challenge.
} 
Furthermore, \bioscan{} contains \textit{lab} images, rather than in situ images like iNat21, diversifying the \textit{image} distribution.

\begin{figure}[t]
    \vskip -6pt
    \small
    \centering
    \includegraphics[width=\columnwidth]{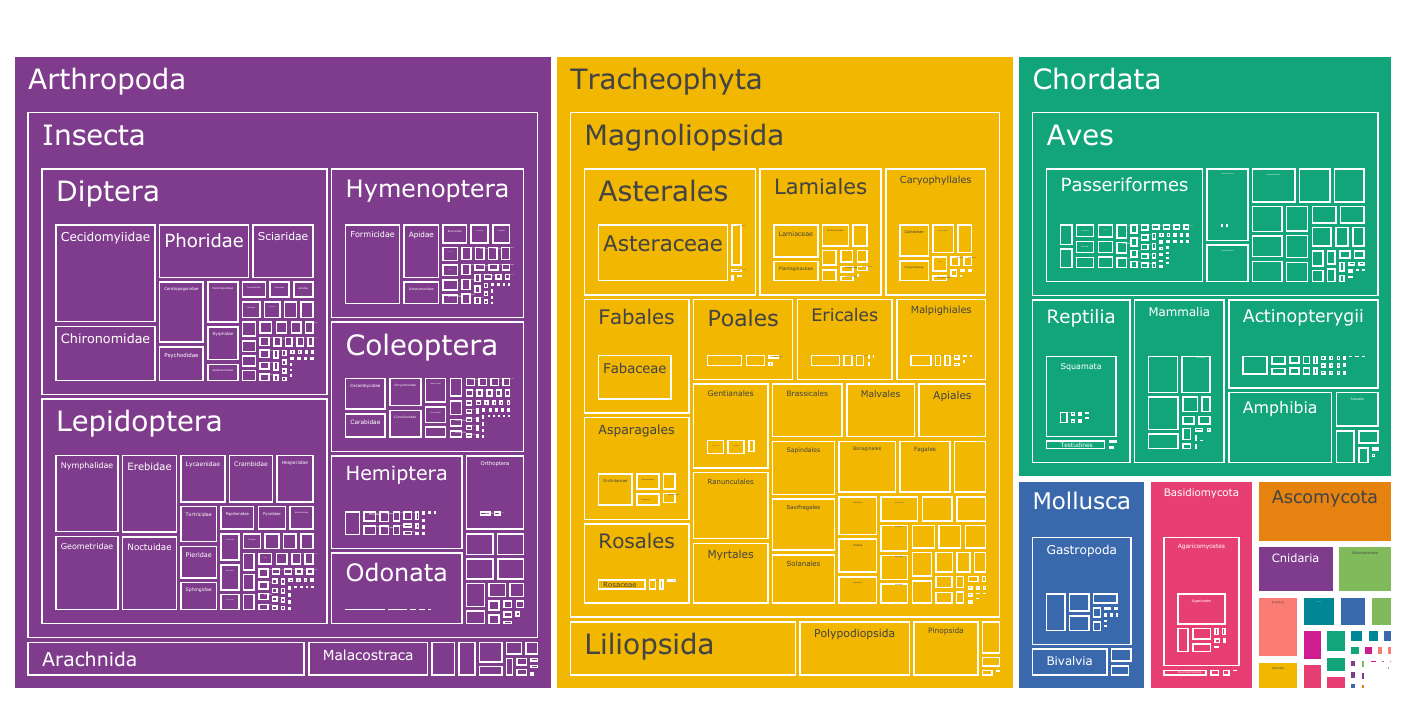}
    \vskip -6pt
    \caption{
        Treemap of the \num{108} phyla in \datasetname{.} 
        Different colors are different phyla; nested boxes represent classes, orders, and families. Box size, not number of inner boxes, represents relative number of samples.
    }
    \label{fig:dataset}
    \vskip -12pt
\end{figure}

\subsection{Metadata \& Aggregation}\label{subsec:metadata}
The 
\datasetname{} dataset integrates iNat21 (training split), our curated EOL dataset, and \bioscan{} by aggregating the images and canonicalizing the labels. 
\emph{This is a highly non-trivial task because taxonomic hierarchies are notoriously noisy and rarely consistent between sources}  \citep{hinchliff2015reconstructing,guralnick2015community,pyle2016towards,patterson2016challenges,rees2017taxonomy}, likely contributing to the prior lack of image datasets large enough to train a foundation-scale vision model for the entire tree of life. 
We carefully unify and backfill taxonomic hierarchies from EOL, the Integrated Taxonomic Information System (ITIS)~\citep{itis2023July}, and iNaturalist with special consideration for the existence of homonyms (genus-species labels shared among higher-order taxa). 
For more information on this process, the challenges, our solutions, and remaining issues, see \cref{app:data-aggregation}.

\subsection{Release \& Statistics}
\cref{tab:training-data} presents dataset statistics: \datasetname{} has over \num{10}M images across more than \num{450}K unique taxonomic names.
\cref{fig:dataset} shows the distribution of images by phyla and the respective lower-rank taxa (order through family).

Our curated training and test datasets (\datasetname{} and \rarespecies{,} described in \cref{subsec:zero-shot}) are available on Hugging Face (with DOIs) under a public domain waiver, to the extent primary source licenses allow. 
This includes CSVs with image metadata and links to the primary sources, accompanied by a GitHub repository with the scripts to generate the datasets.\footnote{
    We encourage future work to cite iNat21 \citep{inat2021}, \bioscan{} \citep{gharaee2023step} and to appropriately attribute images from EOL based on their licenses if citing \datasetname{.} 
}


\begin{table*}[t]
    \small
    \centering
    \setlength\tabcolsep{3pt}
    \renewcommand{\arraystretch}{1.1}
    \scalebox{0.9}{
    \begin{tabularx}{\textwidth}{ccXSSc}
        \toprule
        & \thead{Name} & \thead{Description} & \sithead{Examples} & \sithead{Classes} & \sithead{Labels} \\
        \midrule
        \multirow{4}{*}{\rotatebox{90}{Animals}} & Birds 525 & Scraped dataset of bird images from web search. \citep{piosenka2023birds}  & \num{89885} & \num{525} & Taxonomic \\
        & Plankton & Expert-labeled in situ images of plankton 
        \citep{whoiplankton}. & \num{4080} & \num{102} & Mixed \\
        & Insects & Expert and volunteer-labeled in-the-wild citizen science images of insects \citep{insects}. & \num{4680} & \num{117} & Scientific \\
        & Insects 2 & Mixed common and scientific name classification for insect pests \citep{wu2019insect}. & \num{4080} & \num{102} & Mixed \\
        \cmidrule(lr){2-6}
        \multirow{5}{*}{\rotatebox{90}{Plants \& Fungi}} & PlantNet & Citizen science species-labeled plant images, some drawings \citep{garcin2021plntnetk}. & \num{1000} & \num{25} & Scientific \\
        & Fungi & Expert-labeled images of Danish fungi \citep{picek2022danish}. & \num{1000} & \num{25} & Scientific \\
        & PlantVillage & Museum-style leaf specimens labeled with common names \citep{G2019323}. & \num{1520} & \num{38} & Common \\
        & Medicinal Leaf & Species classification of leaves from mature, healthy medicinal plants \citep{s_j_2020}. & \num{1040} & \num{26} & Scientific \\
        & PlantDoc & 17 diseases for 13 plant species \citep{plntDoc2020}. & \num{1080} & \num{27} & Common \\
        \cmidrule(lr){2-6}
        & \rarespecies{} & Subset of species in the IUCN Red List categories: Near Threatened through Extinct in the Wild (\href{https://www.iucnredlist.org/}{iucnredlist.org}). & \num{12000} & \num{400} & Taxonomic \\
        \bottomrule
    \end{tabularx}
    }
    \vskip -6pt
    \caption{
        Datasets used for evaluation.
        All tasks are classification evaluated with Top-1 accuracy.
    }
    \label{tab:eval-data}
    \vskip -12pt
\end{table*}

\section{Modeling}


\modelname{} is initialized from OpenAI's public CLIP checkpoint and continually pre-trained on \datasetname{} with CLIP's multimodal contrastive learning objective.


\subsection{Why CLIP?}
\label{subsec:why-clip}

Compared with general domain computer vision tasks, one of the most salient differences for the biology domain is its rich label space.
Not only are the taxon labels large in quantity (there are \num{2}M+ recorded species as of 2022 \citep{iucn2022}), but they are also connected with each other in a hierarchical taxonomy. 
This is a challenge for training a good foundation model that can achieve satisfactory coverage and generalization. 
Despite this, the intricate structure in the label space, accumulated through centuries of biology research, provides very rich signal for learning better generalization.
Intuitively, if the label space's structure is successfully encoded in a foundation model, even if the model has not seen a certain species, it will likely have learned a good representation for that species' corresponding genus or family.
Such a hierarchical representation serves as a strong prior to enable few-shot or even zero-shot learning of new taxa.

Many vision foundation models, such as ResNet \citep{he2016deep} and Swin Transformer \citep{liu2021swin}, adopt a supervised classification objective and directly learn the mapping from input images to class indices.
As a result, each class label is treated as a distinct symbol, and their relationships are neglected. 
\emph{A key realization of our work is that the multimodal contrastive learning objective used in CLIP can be repurposed for leveraging the hierarchical structure of the label space.}
This is not an obvious choice; after all, \datasetname{} is largely labeled with class labels and not with free-form text like image captions. 
The autoregressive text encoder naturally embeds the taxonomic hierarchy into a dense label space by conditioning later taxonomic rank representations on higher ranks (\cref{fig:hook}).
While hierarchical classification~\citep{bertinetto2020making,zhang2022use,bjerge2023hierarchical} can also leverage taxonomy, we empirically show that CLIP-style contrastive learning significantly improves generalization (\cref{subsec:design-ablation}).
We note that \textit{repurposing CLIP's multimodal contrastive learning objective for learning hierarchical representations conforming to a taxonomy} is a novel and non-trivial technical contribution.

CLIP trains two uni-modal embedding models, a vision encoder and a text encoder, to (1) maximize feature similarity between \textit{positive} (image, text) pairs and (2) minimize feature similarity between \textit{negative} (image, text) pairs, where positive pairs are from the training data and negative pairs are all other possible (image, text) pairings in a batch.
After training, CLIP's encoder models embed individual instances of their respective modalities into a shared feature space. 
Next, we discuss formatting the text input to CLIP to incorporate the taxonomic structure.

\subsection{Text Types}
\label{subsec:text_types}

\begin{table}[t]
    \small
    \setlength\tabcolsep{3pt}
    \centering
    \begin{tabularx}{\columnwidth}{lX}
        \toprule
        \textbf{Text Type}  & \textbf{Example} \\
        \midrule
        Common & black-billed magpie \\
        Scientific & \textit{Pica hudsonia}\\
        Taxonomic & \textit{Animalia Chordata Aves Passeriformes Corvidae Pica hudsonia}\\
        Scientific + Common & \textit{Pica hudsonia} with common name black-billed magpie \\
        Taxonomic + Common &  \textit{Animalia Chordata Aves Passeriformes Corvidae Pica hudsonia} with common name black-billed magpie \\
        \bottomrule
    \end{tabularx}
    \vskip -6pt
    \caption{
        Text types considered in the training of \modelname{}.
    }
    \label{tab:classname}
    \vskip -12pt
\end{table}

An advantage of CLIP is that the text encoder accepts free-form text. 
In biology, unlike other classification tasks, class names are diversely formatted. 
We consider the following:

\noindent \textbf{Taxonomic name.} A standard seven-level biology taxonomy from higher to lower level is kingdom, phylum, class, order, family, genus and species. For each species, we ``flatten'' the taxonomy by concatenating all labels from root to leaf into a single string, which we call the \textit{taxonomic name}.

\noindent \textbf{Scientific name.} Scientific names are composed of genus and species (e.g., \textit{Pica hudsonia}). 

\noindent \textbf{Common name.} Taxonomy categories are usually Latin, which is not often seen in generalist image-text pre-training datasets.
Instead, the common name, such as ``black-billed magpie,'' is more widespread. 
Note that common names may not have a 1-to-1 mapping to taxa: 
A single species may have multiple common names, or the same common name may refer to multiple species.

For certain downstream use cases of \modelname{}, it might be the case that only one type of label, e.g., scientific names, is available. 
To improve the flexibility at inference time, we propose a \textit{mixed text type} training strategy: at each training step, we pair each input image with a text randomly sampled from all of its available text types (shown in \cref{tab:classname}).
We empirically show that this simple strategy retains the generalization benefits of taxonomic names while providing more flexibility in using other names at inference time (\cref{subsec:text-type-ablation}). 
The final text input to CLIP is the name in the standard CLIP template, e.g., ``a photo of \textit{Pica hudsonia}''. 

\section{Experiments}\label{sec:experiments}

We train \modelname{} on \datasetname{}, compare \modelname{} to general vision models and investigate how our modeling choices affect \modelname{'s} performance.

\subsection{Training and Evaluation Details}\label{subsec:training-details}

To train \modelname{}, we initialize from OpenAI's CLIP weights \citep{radford2021learning} with a ViT-B/16 vision transformer \citep{dosovitskiy2020image} image encoder and a \num{77}-token causal autoregressive transformer text encoder.
We continue pre-training on \datasetname{} for 100 epochs with a cosine learning rate schedule \citep{loshchilov2017sgdr}. 
We train on \num{8} NVIDIA A100-80GB GPUs over \num{2} nodes with a global batch size of \num{32768}. 
We also train a baseline model on only the iNat21 dataset and multiple ablation models on 1M examples randomly sampled from \datasetname{} (\cref{subsec:text-type-ablation,subsec:design-ablation}), following the same procedure for \modelname{} except with a smaller global batch size of \num{16384} on \num{4} NVIDIA A100 GPUs on \num{1} node. 
All hyperparameters and training details are in \cref{app:hyperparameters} and training and evaluation code is publicly available.

We evaluate on \textbf{10 different classification tasks}: the 8 biologically-relevant tasks from \textbf{Meta-Album} \citep{meta-album-2022}, \textbf{Birds 525} \citep{piosenka2023birds} and our new \textbf{\rarespecies{}} task (described in \cref{subsec:zero-shot}).
Meta-Album is a dataset collection for meta-learning, encompassing various subjects.
Specifically, we use the Plankton, Insects, Insects 2, PlantNet, Fungi, PlantVillage, Medicinal Leaf, and PlantDoc datasets.
Our classification tasks cover all four multi-celled kingdoms in the tree of life (animals, plants, fungi, and protists) and have a diverse image distribution (photographs, microscope images, drawings, and museum specimens).
\cref{tab:eval-data} summarizes the datasets; they comprise a variety of label types from full taxonomic names to only scientific or common name.

For \textbf{zero-shot learning}, we follow the same procedure as CLIP. 
For \textbf{few-shot learning}, we follow SimpleShot~\citep{wang2019simpleshot} and use a nearest-centroid classifier. 
For $k$-shot learning, we first randomly sample $k$ examples for each class and obtain the image embedding from the visual encoder of the pre-trained models. 
We then compute the average feature vector of the $k$ embeddings as the centroid for each class. 
All the examples left in the dataset are used for testing. 
After applying mean subtraction and L2-normalization to each centroid and test feature vector, we choose the class with the nearest centroid to the test vector as the prediction. 
We repeat each few-shot experiment \num{5} times with different random seeds and report the mean accuracy in \cref{tab:classification-results}. 
Results with standard deviations are reported in \cref{app:std-dev}.

We compare \modelname{} with the original OpenAI CLIP \citep{radford2021learning} and OpenCLIP \citep{ilharco2021openclip} trained on LAION-400M \citep{schuhmann2021laion400m}.
Intuitively, common names of organisms are most pervasive in the training data of CLIP and OpenCLIP and these models work best with common names.
This is also confirmed in our preliminary tests.
Therefore, we use common names as class labels for CLIP and OpenCLIP by default unless unavailable for a dataset.
\modelname{} can leverage taxonomic names, so we use taxonomic+common names by default. 
However, as noted in \cref{tab:eval-data}, the test datasets come in a variety of labels. 
Whenever the preferred label type is not available, we use labels that come with the dataset.
We also compare to an ImageNet-21K \citep{deng2009imagenet} pre-trained model \citep{steiner2021train} and DINO \citep{caron2021dino} for few-shot classification.





\begin{table*}[t]
    \setlength\tabcolsep{4pt}
    \newcommand{\faded}{\color{Gray}}

    \centering
    \small
    \newcommand{\first}{\bfseries}
    \scalebox{0.93}{
    \begin{tabular}{lSSSSSSSSSSSS}
        \toprule
         & \multicolumn{4}{c}{\thead{Animals}} & \multicolumn{5}{c}{\thead{Plants \& Fungi}} & & & \\  
        \cmidrule(lr){2-5} \cmidrule(lr){6-10} 
        \thead{Model} & \vertical{Birds 525} & \vertical{Plankton} & \vertical{Insects} & \vertical{Insects 2} & \vertical{PlantNet} & \vertical{Fungi} & \vertical{PlantVillage} & \vertical{Med. Leaf} & \vertical{PlantDoc} & \vertical{Rare Species} & \multicolumn{2}{c}{Mean ($\Delta$)} \\
        \midrule
        \faded Random Guessing & \faded 0.2 & \faded 1.2 & \faded 1.0 & \faded 1.0 & \faded 4.0 & \faded 4.0 & \faded 2.6 & \faded 4.0 & \faded 3.7 & \faded 0.3 & \faded 2.2 & \\
        \midrule
        \multicolumn{13}{l}{\textit{Zero-Shot Classification}} \\
        \midrule
        CLIP & 49.9 & 3.2 & 9.1 & 9.8 & 58.5 & 10.2 & 5.4 & 15.9 & 26.1 & 31.8 & 21.9 & \text{--}\\ 
        OpenCLIP & 54.7 & 2.2 & 6.5 & 9.6 & 50.2 & 5.7 & 8.0 & 12.4 & 25.8 & 29.8 & 20.4 & -1.5 \\
        \modelname{} & \first 72.1 & \first 6.1 & \first 34.8 & \first 20.4 & \first 91.4 & 40.7 & \first 24.4 & \first 38.6 & \first 28.4 & \first 38.0 & \first 39.4 & +17.5 \\
        \;\;-- iNat21 Only & 56.1 & 2.6 & 30.7 & 11.5 & 88.2 & \first 43.0 & 18.4 & 25.6 & 20.5 & 21.3 & 31.7 & +9.8 \\
        \midrule
        \multicolumn{13}{l}{\textit{One-Shot Classification}} \\
        \midrule
        CLIP & 43.7 & 25.1 & 21.6 & 13.7 & 42.1 & 17.2 & 49.7 & 70.1 & 24.8 & 28.5 & 33.6 & \text{--} \\
        OpenCLIP & 53.7 & 32.3 & 23.2 & 14.3 & 45.1 & 18.4 & 53.6 & 71.2 & 26.8 &  29.2 & 36.7 & +3.1 \\
        
        Supervised-IN21K & 60.2 & 22.9 & 14.7 & 14.4 & 46.7 & 16.9 & \first 62.3 & 58.6 & 27.7 & 28.0 & 35.2 & +1.6 \\
        DINO & 40.5 & \first 37.0 & 23.5 & 16.4 & 30.7 & 20.0 & 60.0 & 79.2 & 23.7 & 31.0 & 36.2 & +2.6 \\
        
        \modelname{} & 71.8 & 30.6 & \first 57.4 & \first 20.4 & 64.5 & \first 40.3 & 58.8 & \first 84.3 & \first 30.7 & \first 44.9 & \first 50.3 & +16.7 \\
        \;\;-- iNat21 Only & \first 74.8 & 29.6 & 53.9 & 19.7 & \first 67.4 & 35.5 & 55.2 & 75.1 & 27.8 & 36.9 & 47.5 & +13.9 \\
        \midrule
        \multicolumn{13}{l}{\textit{Five-Shot Classification}} \\
        \midrule
        CLIP & 73.5 & 41.2 & 39.9 & 24.6 & 65.2 & 27.9 & 71.8 & 89.7 & 35.2 & 46.0 & 51.5 & \text{--} \\
        OpenCLIP & 81.9 & 52.5 & 42.6 & 25.0 & 68.0 & 30.6 & 77.8 & 91.3 & 42.0 & 47.4 & 55.9 & +4.4 \\
        Supervised-IN21K & 83.9 & 39.2 & 32.0 & 25.4 & 70.9 & 30.9 & \first 82.4 & 82.3 & 44.7 & 47.3 & 53.9 & +2.4 \\
        DINO & 70.8 & \first 56.9 & 46.3 & 28.6 & 50.3 & 34.1 & 82.1 & 94.9 & 40.3 & 50.1 & 55.4 & +3.9 \\
        \modelname{} & 90.0 & 49.3 & \first 77.8 & \first 33.6 & \first 85.6 & \first 62.3 & 80.9 & \first 95.9 & \first 47.5 & \first 65.7 & \first 68.8 & +17.3 \\
        \;\;-- iNat21 Only & \first 90.1 & 48.2 & 73.7 & 32.1 & 84.7 & 55.6 & 77.2 & 93.5 & 41.0 & 55.6 & 65.1 & +13.6 \\
        \bottomrule
    \end{tabular}
    }
    \vskip -8pt
    \caption{
        Zero-, one- and five-shot classification top-1 accuracy for different models.
        \textbf{Bold} indicates best accuracy.
        All models use the same ViT-B/16 architecture.
        ``iNat21 Only'' follows the same procedure as \modelname{} but uses iNat21 instead of \datasetname{.}
        $\Delta$ denotes the difference in mean accuracy with CLIP.
        Supervised-IN21K \citep{steiner2021train} and DINO \citep{caron2021dino} are vision-only models and cannot do zero-shot classification.
    }
    \label{tab:classification-results}
    \vskip -12pt
\end{table*}

\begin{table}[t]
    \newcommand{\first}{\color{RoyalBlue}\bf}
    \renewcommand{\second}{\color{Orange}}
    \small
    \setlength\tabcolsep{1pt}
    \centering
    \begin{tabular}{llSSSSSS}
        \toprule 
        Dataset & Train$\downarrow$Test$\rightarrow$ & \text{Com} & \text{Sci} & \text{Tax} & \text{Sci+Com} & \text{Tax+Com} \\
        \midrule
        \multirow{6}{*}{\datasetshortsmall{}} & Com & \second 24.9 & 9.5 & 10.8 & 22.3 & 21.0 \\
        & Sci & 11.0 & \second 22.3 & 4.5 & 21.5 & 8.0 \\
        & Tax & 11.8 & 10.1 & \second 26.6 & 16.0 & 24.8 \\
        & Sci+Com & 24.5 & 12.9 & 12.6 & \second 28.0 & 24.9 \\
        & Tax+Com & 20.5 & 8.0 & 19.7 & 24.0 & \second 30.4 \\
        & Mixture & \first 26.1 & \first 24.9 & \first 26.7 & \first 29.5  & \first 30.9 \\
        \midrule
        iNat21-2.7M & Mixture & 20.4 & 14.7 & 15.6 & 20.9 & 21.3 \\
        \datasetshortname{} & Mixture & 31.6 & 30.1 & 34.1 & 37.0 & 38.0 \\
        \bottomrule
    \end{tabular}
    \vskip -6pt
    \caption{
        Zero-shot accuracy on species not seen during training (\rarespecies{} task).
        Different rows and columns indicate different text types during training and test time, respectively.
        {\color{RoyalBlue}\bf Blue} indicates best accuracy and {\color{Orange}Orange} indicates second-best. 
        Using the taxonomic name over the scientific name always improves accuracy ({\color{Orange}\num{22.3}}$\rightarrow${\color{Orange}\num{26.6}} and {\color{Orange}\num{28.0}}$\rightarrow${\color{Orange}\num{30.4}}).
        The final rows use the full iNat21 dataset and \datasetname{} for reference.
    }
    \label{tab:text-type-ablation}
    \vskip -18pt
\end{table}

\subsection{Can \modelname{} Generalize to Unseen Taxa?}\label{subsec:zero-shot}
Taxonomic names are added, removed, and changed as biologists discover and classify new and existing species.
\modelname{} should generalize to unseen taxonomic names to avoid re-training for every new species.
To empirically answer whether \modelname{} generalizes well to unseen taxa, we introduce a new evaluation task that is both biologically and empirically motivated: \textbf{\rarespecies{.}}

Classifying ``rare'' species is an important and challenging computer vision application in biology, particularly in the context of global conservation efforts \citep{tuia2022perspectives}. 
To the best of our knowledge, there is no diverse, publicly available rare species classification dataset. 
Recently published work \citep{Mou_Liang_Hu_Meng_Han_Xu_2023,liu2023lote} lack species diversity with only a dozen classes.
We aim to fill this gap, collecting all $\approx25$K species on the IUCN Red List (\href{https://www.iucnredlist.org/}{iucnredlist.org}) classified\footnote{
    IUCN has classified \num{150388} species and generally updates their list twice per year (\href{https://www.iucnredlist.org/assessment/updates}{IUCN Update Schedule}). 
    The classifications used for this dataset are current as of July 13, 2023.
} as Near Threatened, Vulnerable, Endangered, Critically Endangered, or Extinct in the Wild. 
We select \num{400} such species
represented by at least \num{30} images in our EOL dataset, then remove them
from \datasetname{,} creating an \textit{unseen} \rarespecies{} test set with \num{30} images per species. 
This dataset demonstrates (1) \modelname{'s} out-of-distribution generalization to unseen taxa, (2) \modelname{'s} potential applications, and (3) provides a crucial dataset for the community to address the ongoing biodiversity crisis.

\noindent \textbf{Results}.
\cref{tab:classification-results} shows that \modelname{} substantially outperforms both baseline CLIP models, as well as the iNat21-trained CLIP model, at zero-shot classification, especially on unseen taxa (see the ``Rare Species'' column).
We attribute \modelname{'s} strong zero-shot performance on this broad and diverse set of tasks to the broad and diverse classes in \datasetname{.}
We explore how data diversity leads to broadly useful image representations in \cref{subsec:text-type-ablation}.

\subsection{How Do Text Types Affect Generalization?}\label{subsec:text-type-ablation}

We investigate how different text types affect zero-shot generalization by training \modelname{} on a 10\% subset of \datasetname{} (10\% due to computational constraints).
We use our Rare Species dataset because the test classes have every text type, and all species are excluded from training, making it ideal for testing generalization to unseen taxa.
Prior works find that the diversity of captions makes stronger vision models \citep{nguyen2022quality} and randomly use one of five different captions for each image during training rather than a single fixed caption \citep{santurkar2022caption}.
Similarly, we use a mixed text type strategy (\cref{subsec:text_types}).
How does that affect performance?

\noindent \textbf{Results}. The zero-shot ablation results are in \cref{tab:text-type-ablation}; there are several salient observations. 
First, using taxonomic+common names yields the strongest performance, showing the importance of incorporating the taxonomic structure for generalization.
Second, when only using a single text type for training, performance degrades substantially when a different text type is used at test time. 
Using mixed text types for training yields consistently strong performance across all text types during testing. 
These results indicate that mixed text type pre-training largely retains the generalization benefits of using taxonomic names while also providing flexibility of different text types for inference, an important property for a foundation model that may be used for diverse downstream tasks.
Finally, using 1M examples from \datasetname{} outperforms using 2.7M examples from iNat21, further confirming the importance of the added data diversity from \datasetname{.}


\begin{figure*}[t]
    \small
    \centering
    \vskip -12pt
    \includegraphics[width=\textwidth]{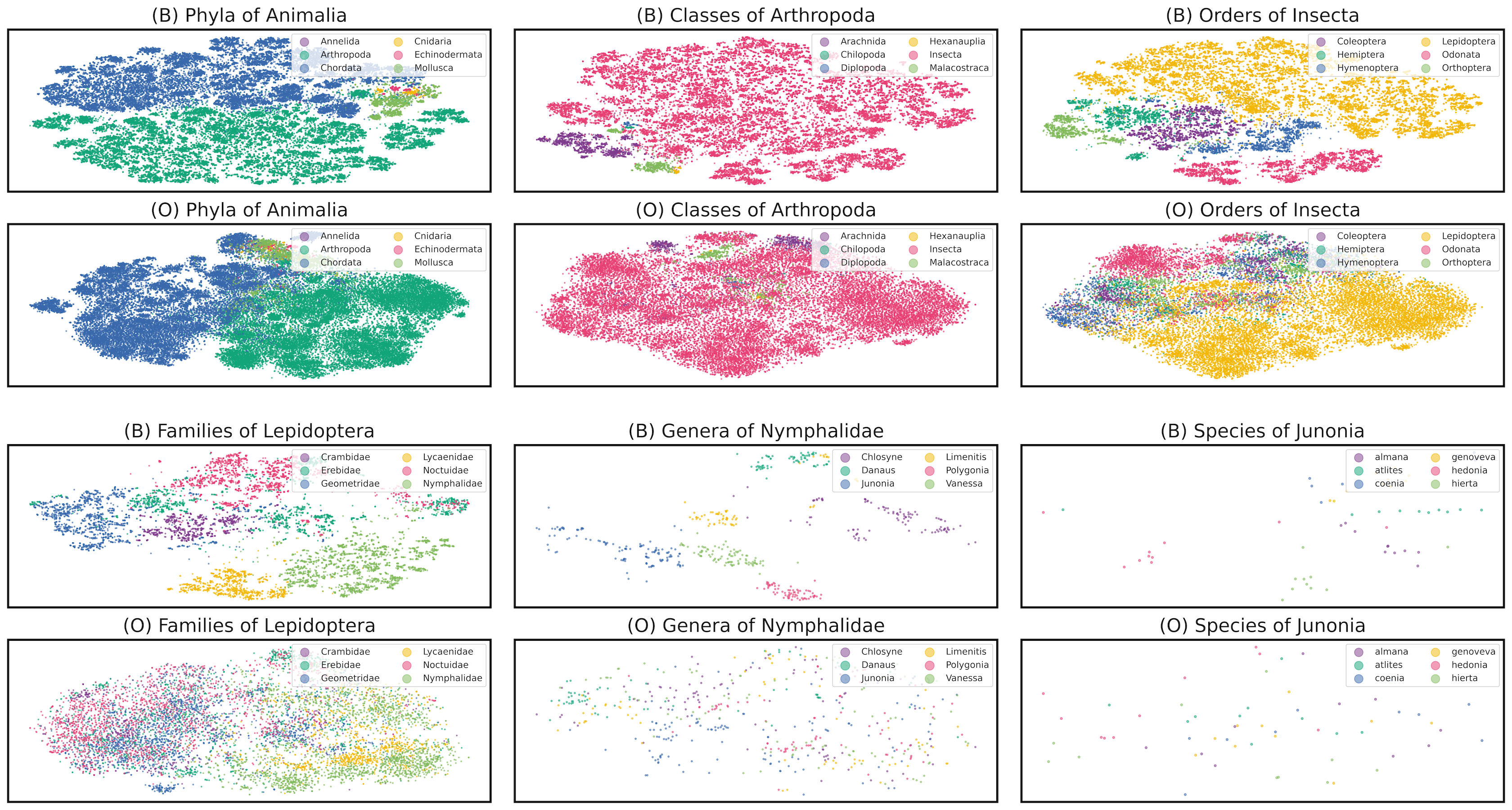}
    \vskip -6pt
    \caption{
        T-SNE visualization of image features, colored by taxonomic labels. 
        \modelname{} (B) is visualized in the first and third row and OpenAI's CLIP (O) is visualized in the second and fourth rows.
        \modelname{'s} features better preserve the hierarchical structure: while both \modelname{} and CLIP cleanly separate the phyla in the Animalia Kingdom (top left), only \modelname{} successfully separates the orders in the Insecta Class (top right) and the families in the Lepidoptera Order (bottom left).
    }
    \label{fig:intrinsic}
    \vskip -12pt
\end{figure*}

\begin{table}[t]
    \centering
    \small
    \newcommand{\first}{\bfseries}
    \scalebox{0.9}{
    \begin{tabular}{lSS}
        \toprule
        \thead{Objective} & \sithead{Mean 1-Shot} & \sithead{Mean 5-shot} \\
        \midrule
        Cross-entropy & 16.5 & 26.2 \\
        Hier. cross-entropy & 19.3 & 30.5 \\
        CLIP & \first 44.7 & \first 63.8 \\ 
        \bottomrule
    \end{tabular}
    }
    \vskip -6pt
    \caption{
        One- and five-shot classification top-1 accuracy for different pre-training objectives on \datasetsmall{.} Results are macro-averaged over all the test sets in \cref{tab:classification-results}.
    }
    \label{tab:objective-ablation-results}
    \vskip -12pt
\end{table}

\subsection{Is the CLIP Objective Necessary?}\label{subsec:design-ablation}

Using the CLIP objective to pre-train on a labeled image dataset is an unintuitive decision (\citet{goyal2023finetune} fine-tune using the CLIP objective, but do not pretrain).
We justify our choice by training two ViT-B/16 models on \datasetsmall{} using a cross-entropy classification loss and a multitask hierarchical variant, then compare them against the CLIP objective under the few-shot setting. 
The multitask hierarchical training objective is to predict the labels for kingdom, phylum, etc., down to species, using cross entropy for each level of the taxonomy, then summing those losses \citep{bjerge2023hierarchical}.
Pseudo-code is provided in \cref{alg:hierarhical-multitask}.

\noindent \textbf{Results}.
We evaluate each model on the same set of \num{10} tasks but only in the one-shot and five-shot settings because non-CLIP models cannot do zero-shot classification.
We report mean accuracy in \cref{tab:objective-ablation-results}.
The hierarchical classification model outperforms simple classification and is comparable to the CLIP baseline (see \cref{tab:classification-results}).
However, the CLIP objective massively outperforms both baselines and strongly justifies our repurposing of the CLIP objective.

\subsection{Can \modelname{} Classify More Than Species?}\label{subsec:few-shot}

\modelname{} is trained on a (contrastive) species-classification objective, which might limit its use beyond species classification.
We consider plant diagnosis with the PlantVillage and PlantDoc datasets, which require classifying both species and disease (if any).
Large-scale data labeling is expensive, but biologists always label several instances for field guides or museum collections.
Few-shot classification is thus an ideal setting for this sort of task transfer.

\noindent \textbf{Results}.
\modelname{} outperforms baselines at classifying plant diseases based on visual symptoms, in both zero-shot and few-shot settings (see PlantVillage and PlantDoc in \cref{tab:classification-results}).
While \citet{radford2021learning} find that CLIP one-shot and two-shot classification is often worse than zero-shot (because few-shot settings cannot use the semantic information in the class name), \modelname{} has learned useful visual representations that are useful even with only one labeled example: \modelname{'s} mean one-shot accuracy is 9.1\% higher than zero-shot accuracy.




\subsection{Does \modelname{} Learn the Hierarchy?}\label{subsec:intrinsic-eval}

\modelname{} demonstrates strong performance in a low-data regime on our extrinsic evaluation, but why? 
We further conduct an intrinsic evaluation and visualize \modelname{'s} learned image representations to shed light on this question (\cref{fig:intrinsic}). 
We embed image representations from iNat21's validation set (unseen during training) using t-SNE \citep{van2008visualizing} and color the points by the image's taxonomic label. 
For each plot, we run t-SNE independently on the subset of examples under the labeled taxonomical rank.
Each plot visualizes one taxonomic hierarchy rank and the top six categories of the next rank, e.g., the top left plot visualizes the six most common phyla in the Animalia kingdom.
At higher ranks like kingdom (omitted for space) and phylum, both CLIP and \modelname{} have good separation, but \modelname{’s} representations are more fine-grained and contain a richer clustering structure. 
At lower ranks, \modelname{} produces evidently more separable features, while CLIP’s features are cluttered and lack a clear structure. 
\cref{app:example-predictions} has more qualitative results and visuals.


\section{Related Work}

\noindent \textbf{Multimodal foundation model training data}.
CLIP \citep{radford2021learning} trained state-of-the-art vision models from noisy, web-scale (\num{100}M+) image-text datasets using a contrastive objective that is optimized for image retrieval.
ALIGN \citep{jia2021scaling} and BASIC \citep{pham2023combined} further scaled the number of training examples from 400M to 6.6B, improving vision representation quality.
However, further work \citep{fang2022data, nguyen2022quality, gadre2023datacomp, xu2023cit, xu2023demystifying} all find that \emph{dataset diversity and better alignment between the image and caption semantics} are more important than dataset size and lead to stronger performance on downstream tasks. 
\emph{\datasetname{} emphasizes the importance of diversity}, adding over \num{440}K classes
to iNat21's \num{10}K and leading to stronger zero-shot performance. 

\noindent \textbf{Domain-specific CLIPs}.
Domain-specific training often beats general training \citep{gu2021domain, chia2022fashionclip}, but 
subject-matter experts are often too expensive to hire to label large-scael domain-specific datasets.
Image-text training is thus particularly potent because models can learn from noisy image-text pairs.
\citet{ikezogwo2023quilt} and \citet{lu2023towards} gathered \num{1}M+ image-text pairs for computational pathology. 
We gather $10\times$ the images, emphasizing class diversity.

\noindent \textbf{Hierarchy in computer vision}.
Hierarchy in computer vision is well-studied, in part because ImageNet \citep{russakovsky2015imagenet} classes are from the hierarchical WordNet \citep{miller1995wordnet}.
\citet{bilal2017convolutional} study model predictions on ImageNet and find that model confusion patterns follow the hierarchical class structure.
They incorporate hierarchy into AlexNet's architecture \citep{krizhevsky2012imagenet} and improve ImageNet top-1 error by \num{8}\% absolute.
\citet{bertinetto2020making} measure image classifiers' mistake severity and propose alternative objectives that incorporate hierarchy, reducing mistake severity at the expense of worsening top-1 accuracy. 
\citet{zhang2022use} propose a contrastive objective where the hierarchical distance between labels corresponds to the desired distance in the embedding space, and outperform cross-entropy on ImageNet and iNat17 \citep{inat2017}.
We apply hierarchical classification to \num{454}K unique classes through a repurposed CLIP objective, while prior work applied hierarchies to smaller label spaces.


\noindent \textbf{Computer vision for biology}.
Fine-grained classification is a classic challenge in computer vision, and biological images are often used to benchmark models.
\citet{wah2011cub, berg2014birdsnap, piosenka2023birds} all use bird species classification to evaluate fine-grained classification ability.
Biology tasks are used for contrastive learning frameworks \citep{xiao2021what,cole2022does}, weakly supervised object detection \citep{cole2022label} and semi-supervised learning methods \citep{he2024species196}.

\section{Conclusion}

We introduce \datasetname{} and \modelname{}, a large-scale diverse biology image dataset and a foundation model for the tree of life, respectively.
Through extensive evaluation, we show that \modelname{} is a strong fine-grained classifier for biology in both zero- and few-shot settings.
We corroborate our hypothesis that using the entire taxonomic name leads to stronger generalization than other caption types through an ablation on unseen species and by visualizing \modelname{'s} representations, finding that \modelname{-embedded} images better match the taxonomic hierarchy.

Although the CLIP objective efficiently learns visual representations over \num{450}K taxa, \modelname{} is fundamentally trained to do classification.
Future work will further scale up the data, e.g., incorporating more than \num{100}M research-grade images from \href{https://inaturalist.org}{iNaturalist}, and collect richer textual descriptions of species' appearances such that \modelname{} can extract fine-grained trait-level representations. 

\section*{Acknowledgements}

We thank the \href{https://imageomics.osu.edu/about/team}{Imageomics team} (including Josef Uyeda, Jim Balhoff, Dan Rubenstein, Hank Bart, Hilmar Lapp, Sara Beery and Dipanjyoti Paul) and the OSU NLP group for their valuable feedback, the \bioscan{} and iNaturalist teams for sharing their data, and Jennifer Hammock at EOL for her help accessing EOL's images.
Our research is supported by NSF OAC 2118240 and resources from the Ohio Supercomputer Center \citep{OhioSupercomputerCenter1987}.

{
    \small
    \bibliographystyle{ieeenat_fullname}
    \bibliography{bib,tanya}
}

\clearpage
\appendix

\setcounter{table}{0}
\renewcommand\thetable{\Alph{section}\arabic{table}}
\setcounter{figure}{0}
\renewcommand\thefigure{\Alph{section}\arabic{figure}}

\section*{Appendices}

Many details are omitted in the main text because of space concerns; we present relevant details here.
\begin{enumerate}[nosep]
    \item{\cref{app:reproduce}: Reproducibility statement}
    \item{\cref{app:ethics}: Ethics statement}
    \item{\cref{app:data-aggregation}: Details of training data aggregation}
    \item{\cref{app:hyperparameters}: Training details and hyperparameters}
    \item{\cref{app:std-dev}: Standard deviations for few-shot results}
    \item{\cref{app:example-predictions}: Example zero-shot predictions on our evaluation tasks.}
    \item{\cref{app:more-text-type}: Additional text-type results}
    \item{\cref{app:gzsl}: Generalized zero-shot learning setting}
\end{enumerate}

\section{Reproducibility Statement}\label{app:reproduce}

We ensure reproducibility of our results by releasing our datasets (\datasetname{} and \rarespecies{}), data pre-processing code, training code, evaluation code, code to generate all figures (\cref{fig:dataset,fig:intrinsic}), and pre-trained model weights.
With these resources, anyone with sufficient compute resources can download the original data, then reproduce the pre-processing, training, and evaluation.
For those with limited compute, the pre-trained model weights enable full reproducibility of our evaluation results (\cref{sec:experiments}).

We provide DOIs as permanent references to our new digital assets:
\begin{itemize}
    \item{\datasetname{}: \href{https://doi.org/10.57967/hf/1972}{doi:10.57967/hf/1972}}
    \item{\rarespecies{}: \href{https://doi.org/10.57967/hf/1981}{doi:10.57967/hf/1981}}
    \item{\modelname{}: \href{https://doi.org/10.57967/hf/1511}{doi:10.57967/hf/1511}}
    \item{Code: \href{https://doi.org/10.5281/zenodo.10895871}{doi:10.5281/zenodo.10895871}}
\end{itemize}

\section{Ethics Statement}\label{app:ethics}

We are not aware of any major ethical issues that arise from our work. 
\modelname{} is further pre-trained from the original CLIP model; many of the same concerns (class design, surveillance, etc.) apply; however, these concerns are discussed in great detail in \citet{radford2021learning},
so we will focus on addressing these concerns as they relate to the biological addition provided in \modelname{}. 

How could \modelname{} affect endangered species--does \modelname{} or \datasetname{} pose a threat by aiding poachers?
Though \modelname{} leads to improved automatic species classification, it does not include specific geographic information such as GPS coordinates.
Furthermore, animal conservation status is not included during training.

Could \modelname{} have a negative impact on biologists?
\modelname{} is designed to combine visual cues with an established taxonomic hierarchy to aid in scientific discovery. 
Concerns regarding over-reliance on model predictions is a warning that accompanies many--if not all--contemporary models and is not unique to \modelname{.}
The goal is for \modelname{} to aid biologists in their work, not to replace them. 
As such, it is important for users to retain that understanding/context when applying \modelname{} to downstream tasks.

\section{Training Data Aggregation}\label{app:data-aggregation}
We aggregate images and labels from the iNat21 training data, \bioscan{'s}, and data downloaded from \href{https://eol.org}{EOL}. 
While every image has at least one taxonomic rank labeled, full taxonomic hierarchies and common names are scraped on a best-effort basis from metadata providers, including iNaturalist
(\href{https://www.inaturalist.org/pages/developers}{iNaturalist Taxonomy DarwinCore Archive}), Encyclopedia of Life (\href{https://opendata.eol.org/dataset/tram-807-808-809-810-dh-v1-1/resource/942b7420-4f44-4c11-aad9-bd99a31f12ba}{eol.org}) and Integrated Taxonomic Information System (ITIS) (\href{https://www.itis.gov/}{itis.gov}). 

We create a lookup between scientific name and taxonomic hierarchy and a lookup between scientific name and common name.  
We populate these lookups using the following sources in order of descending prioritization, as earlier sources are considered more authoritative. That is, if a duplicate appears in a later source, it is superseded by the higher priority source: \bioscan{} metadata, \href{https://opendata.eol.org/dataset?organization=encyclopedia_of_life}{EOL aggregate datasets}: information retrieved using EOL page IDs with the \href{https://eol.org/docs/what-is-eol/classic-apis}{pages API}, which checks for a match in the ITIS hierarchy for higher-level taxa standardization (setting aside homonyms for proper linkage). The full list of taxa and vernacular names provided by iNaturalist and the iNat21 training set class names were maintained. From here, any taxa that could not be resolved using these sources were fed through the \href{https://resolver.globalnames.org/api}{Global Names Resolver (GNR) API}. Overall we were able to achieve 84\% full taxa labeling for images in \datasetname{}, for context, 10\% of \datasetname{} is only labeled down to the family rank (\bioscan{}), thus, genus-species information is not available. 

Despite our efforts, we discovered after training that some hemihomonyms were mislabeled at higher-level taxa (family up to kingdom).
This impacts approximately $0.1-0.2\%$ of our data. 
We are in the process of developing a more robust solution to taxonomic labeling which will also account for re-naming (as is currently in process for many bird species). 
We intend to release a patch alongside the solution.

\section{Hyperparameters \& Training Details}\label{app:hyperparameters}

\cref{tab:common-hyperparams,tab:specific-hyperparams} contain our training hyperparameters for the different models.
\cref{tab:specific-hyperparams} notes the different epochs at which we had the lowest validation loss, as evaluated using the CLIP objective on the validation split of \datasetname{} (even for the \datasetsmall{} models).
We will release our training code upon acceptance.

\begin{table}[t]
    \centering
    \small
    \begin{tabular}{lc}
        \toprule
        Hyperparameter & Value \\
        \midrule
        Architecture & ViT-B/16 \\
        Max learning rate & $1\times10^{-4}$ \\
        Warm-up steps & \num{1000} \\
        Weight Decay & \num{0.2} \\
        Input Res. & $224\times224$ \\
        \bottomrule
    \end{tabular}
    \caption{Common hyperparameters among all models we train.}
    \label{tab:common-hyperparams}
\end{table}

\begin{table}[t]
    \centering
    \small
    \begin{tabular}{lccc}
        \toprule
        Dataset & Text Type & Batch Size & Epoch \\
        \midrule
        \datasetname{} & Mixture & 32K & 100 \\
        iNat21 Only & Mixture & 16K & 65\\ 
        \midrule
        \multirow{6}{*}{\datasetsmall{}} & Common & \multirow{6}{*}{16K} & 86 \\ 
        & Scientific & & 87\\ 
        & Taxonomy & & 87 \\
        & Sci+Com & & 87 \\
        & Tax+Com & & 86 \\
        & Mixture & & 91 \\
        \bottomrule
    \end{tabular}
    \caption{
        Hyperparameters that differ between the various models we train.
        We use a smaller batch size and only 1M examples for our text-type ablation because of limited compute.
    }
    \label{tab:specific-hyperparams}
\end{table}

We trained a hierarchical classification model in \cref{subsec:design-ablation}.
Python pseudocode for the training objective is in \cref{alg:hierarhical-multitask}.
We will publicly release full training code upon acceptance.

\begin{listing}[t]
\footnotesize
\begin{minted}{python}
import torch.nn.functional as F

def forward(vit, heads, images, h_labels):
  """
  vit: vision transformer.
  heads: linear layers, one for each taxonomic 
         rank.
  images: batch of input images
  h_labels: hierarchical labels; each image has 
            7 labels
  """
  img_feats = vit(images)
  h_logits = [head(img_feats) for head in heads]
  losses = [F.cross_entropy(logits, label) 
    for logits, labels in zip(h_logits, h_labels)]
  return sum(losses)
\end{minted}
\caption{
    Python code to calculate the hierarchical multitask objective. Each image has 7 class labels: one for each taxonomic rank. 
    The ViT calculates dense features for each image, then each taxonomic rank has its own linear layer that produces logits. 
    By summing the losses, the ViT learns to produce image features that are useful for classifying images at multiple taxonomic ranks.
}
\vskip -8pt
\label{alg:hierarhical-multitask}
\end{listing}

\section{Standard Deviation of Main Results}
\label{app:std-dev}

\cref{tab:std1,tab:std2} show the accuracy with standard deviation over five runs on the test sets presented in \cref{tab:eval-data}.
Since we randomly select the training examples from the datasets for few-shot, accuracies vary based on which examples are train examples and which are test examples.
However, the variation is small enough that our conclusions in \cref{subsec:few-shot} still hold.
Zero-shot results are deterministic and have no variation.

\begin{table*}[ht]
    \centering
    \small
    \begin{tabular}{lccccc}
        \toprule
        \thead{Model} & \vertical{Birds 525} & \vertical{Plankton} & \vertical{Insects} & \vertical{Insects 2} & \vertical{Rare Species} \\
        \midrule
        \multicolumn{6}{l}{\textit{One-Shot Classification}} \\
        \midrule
        CLIP & $43.7 \pm 0.26$ & $25.1 \pm 0.71$ & $21.6 \pm 1.05$ & $13.7 \pm 1.09$ & $28.5 \pm 0.65$\\
        OpenCLIP &$53.7 \pm 0.52$ & $32.3 \pm 0.63$ & $23.2 \pm 1.58$ & $14.3 \pm 0.67$ & $ 29.2 \pm 0.64$ \\
        Supervised-IN21K & $60.2 \pm 1.02$ & $22.9 \pm 0.84$ & $14.7 \pm 1.38 $ & $14.4 \pm 0.90$ & $28.0 \pm 0.77$\\
        DINO & $40.5 \pm 0.96$ & $\mathbf{37.0 \pm 1.39}$ & $23.5 \pm 1.49$ & $16.4 \pm 0.78$ & $31.0 \pm 0.89$\\
        \modelname{} &$71.8 \pm 0.47$ & $30.6 \pm 0.77$ & $\mathbf{57.4 \pm 2.4}$ & $\mathbf{20.4 \pm 1.28}$ & $\mathbf{44.9 \pm 0.73}$\\
        \;\;-- iNat21 Only & $\mathbf{74.8 \pm 0.89}$ & $29.6 \pm 0.82$ & $53.9 \pm 0.97$ & $19.7 \pm 0.80$ & $36.9 \pm 1.02$\\
        \midrule
        \multicolumn{6}{l}{\textit{Five-Shot Classification}} \\
        \midrule
        CLIP & $73.5 \pm 0.37$ & $41.2 \pm 1.01$ & $39.9 \pm 0.86$ & $24.6 \pm 0.90$ & $46.0 \pm 0.33$\\
        OpenCLIP & $81.9 \pm 0.25$ & $52.5 \pm 0.83$ & $42.6 \pm 0.82$ & $25.0 \pm 0.83$ & $47.4 \pm 0.34$\\
        Supervised-IN21K & $83.9 \pm 0.15$ & $39.2 \pm 1.66$ & $32.0 \pm 1.90$ & $25.4 \pm 2.13$ & $47.3 \pm 0.41$ \\
        DINO & $70.9 \pm 0.34$ & $\mathbf{56.9 \pm 1.61}$ & $46.3 \pm 1.37$ & $28.6 \pm 1.59$ & $50.1 \pm 0.47$ \\
        \modelname{} & $90.0 \pm 0.12$ & $49.3 \pm 1.14$ & $\mathbf{77.8 \pm 0.81}$ & $\mathbf{33.6 \pm 0.74}$ & $\mathbf{65.7 \pm 0.43}$ \\
        \;\;-- iNat21 Only & $\mathbf{90.1 \pm 0.08}$ & $48.2 \pm 1.24$ & $73.7 \pm 0.65$ & $32.1 \pm 1.97$ & $55.6 \pm 0.16$\\
        \bottomrule
    \end{tabular}
    \caption{Accuracy with standard deviation of five runs on animals and rare species in \cref{tab:classification-results}}
    \label{tab:std1}
\end{table*}

\begin{table*}[ht]
    \centering
    \small
    \begin{tabular}{lccccc}
        \toprule
        \thead{Model} & \vertical{PlantNet} & \vertical{Fungi} & \vertical{PlantVillage} & \vertical{Med. Leaf} & \vertical{PlantDoc} \\
        \midrule
        \multicolumn{6}{l}{\textit{One-Shot Classification}} \\
        \midrule
        CLIP & $42.1 \pm 3.40$ & $17.2 \pm 0.78$ & $49.7 \pm 2.53$ & $70.1 \pm 2.83$ & $24.8 \pm 1.61$\\
        OpenCLIP & $45.1 \pm 3.40$ & $18.4 \pm 1.26$ & $53.6 \pm 0.79$ & $71.2 \pm 3.58$ & $26.8 \pm 1.45$\\
        Supervised-IN21K & $46.7 \pm 6.30$ & $16.9 \pm 2.32$ & $\mathbf{62.3 \pm 2.28}$ & $58.6 \pm 4.45$ & $27.7 \pm 2.86$ \\
        DINO & $30.7 \pm 3.79$ & $20.0 \pm 1.53$ & $60.0 \pm 2.15$ & $79.2 \pm 2.74$ & $23.7 \pm 2.48$ \\
        \modelname{} & $64.5 \pm 2.15$ & $\mathbf{40.3 \pm 3.00}$ & $58.8 \pm 2.83$ & $\mathbf{84.3 \pm 1.90}$ & $\mathbf{30.7 \pm 1.75}$  \\
        \;\;-- iNat21 Only & $\mathbf{67.4 \pm 4.54}$ & $35.5 \pm 2.93$ & $55.2 \pm 1.58$ & $75.1 \pm 1.16$ & $27.8 \pm 1.31$\\
        \midrule
        \multicolumn{6}{l}{\textit{Five-Shot Classification}} \\
        \midrule
        CLIP & $65.2 \pm 1.25$ & $27.9 \pm 2.54$ & $71.8 \pm 1.46$ & $89.7 \pm 1.45$ & $35.2 \pm 1.59$\\
        OpenCLIP & $68.0 \pm 0.86$ & $30.6 \pm 1.26$ & $77.8 \pm 1.28$ & $91.3 \pm 0.85$ & $42.0 \pm 1.32$\\
        Supervised-IN21K & $70.9 \pm 2.45$ & $30.9 \pm 2.64$ & $\mathbf{82.4 \pm 1.53}$ & $82.3 \pm 3.81$ & $44.7 \pm 2.26$ \\
        DINO & $50.3 \pm 3.20$ & $34.1 \pm 2.87$ & $82.1 \pm 1.31$ & $94.9 \pm 1.30$ & $40.3 \pm 2.32$ \\
        \modelname{} & $\mathbf{85.6 \pm 1.79}$ & $\mathbf{62.3 \pm 1.82}$ & $80.9 \pm 1.04$ & $\mathbf{95.9 \pm 1.07}$ & $\mathbf{47.5 \pm 1.35}$ \\
        \;\;-- iNat21 Only & $84.7 \pm 1.24$ & $55.6 \pm 2.61$ & $77.2 \pm 0.68$ & $93.5 \pm 1.13$ & $41.0 \pm 1.75$\\
        \bottomrule
    \end{tabular}
    \caption{Accuracy with standard deviation of five runs on plants and fungi in \cref{tab:classification-results}}
    \label{tab:std2}
\end{table*}

\section{Example Predictions}\label{app:example-predictions}

\begin{figure*}[t]
    \begin{subfigure}[b]{0.33\textwidth}
        \includegraphics[width=\textwidth]{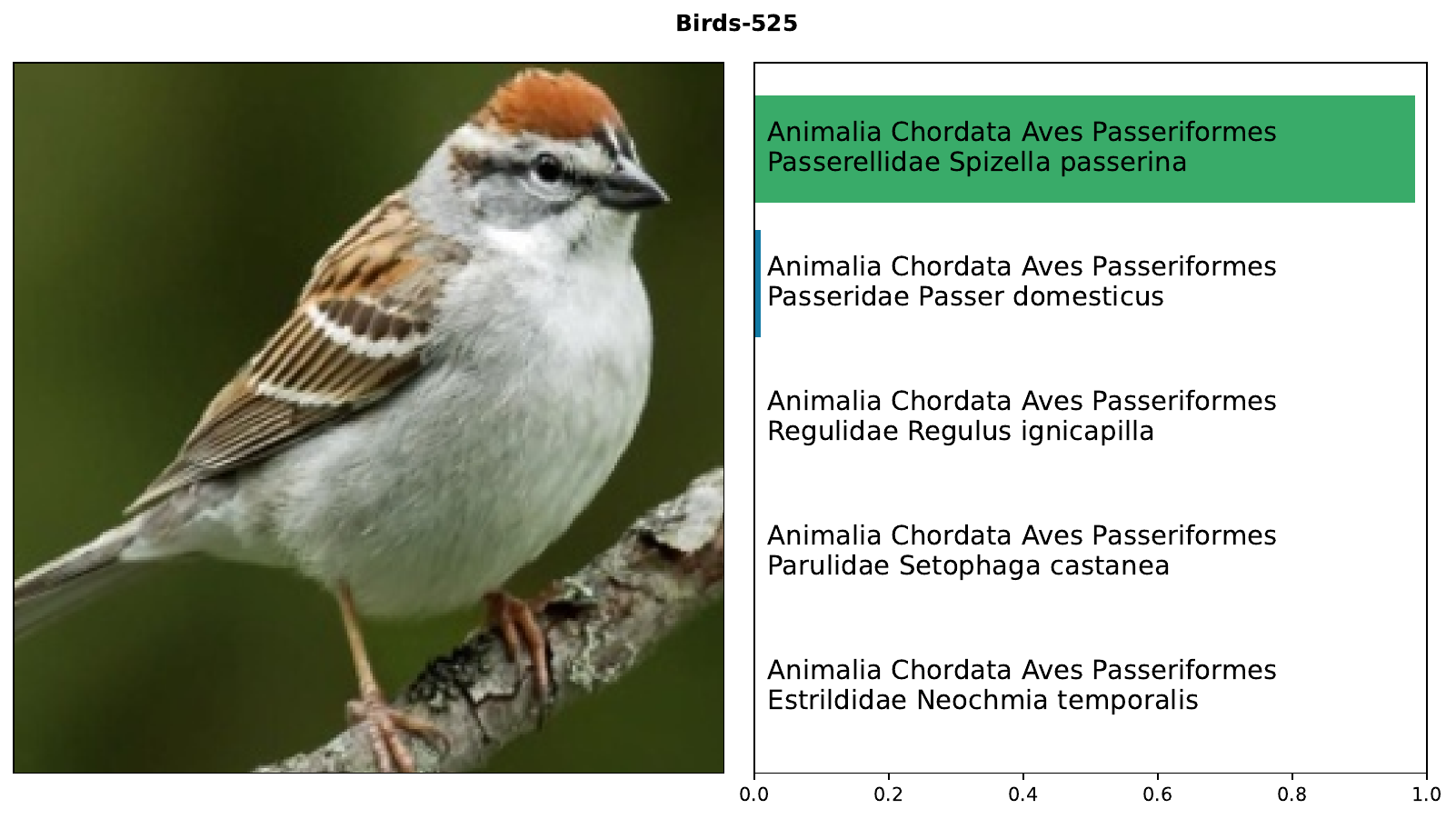}
    \end{subfigure}
    \hfill
    \begin{subfigure}[b]{0.33\textwidth}
        \includegraphics[width=\textwidth]{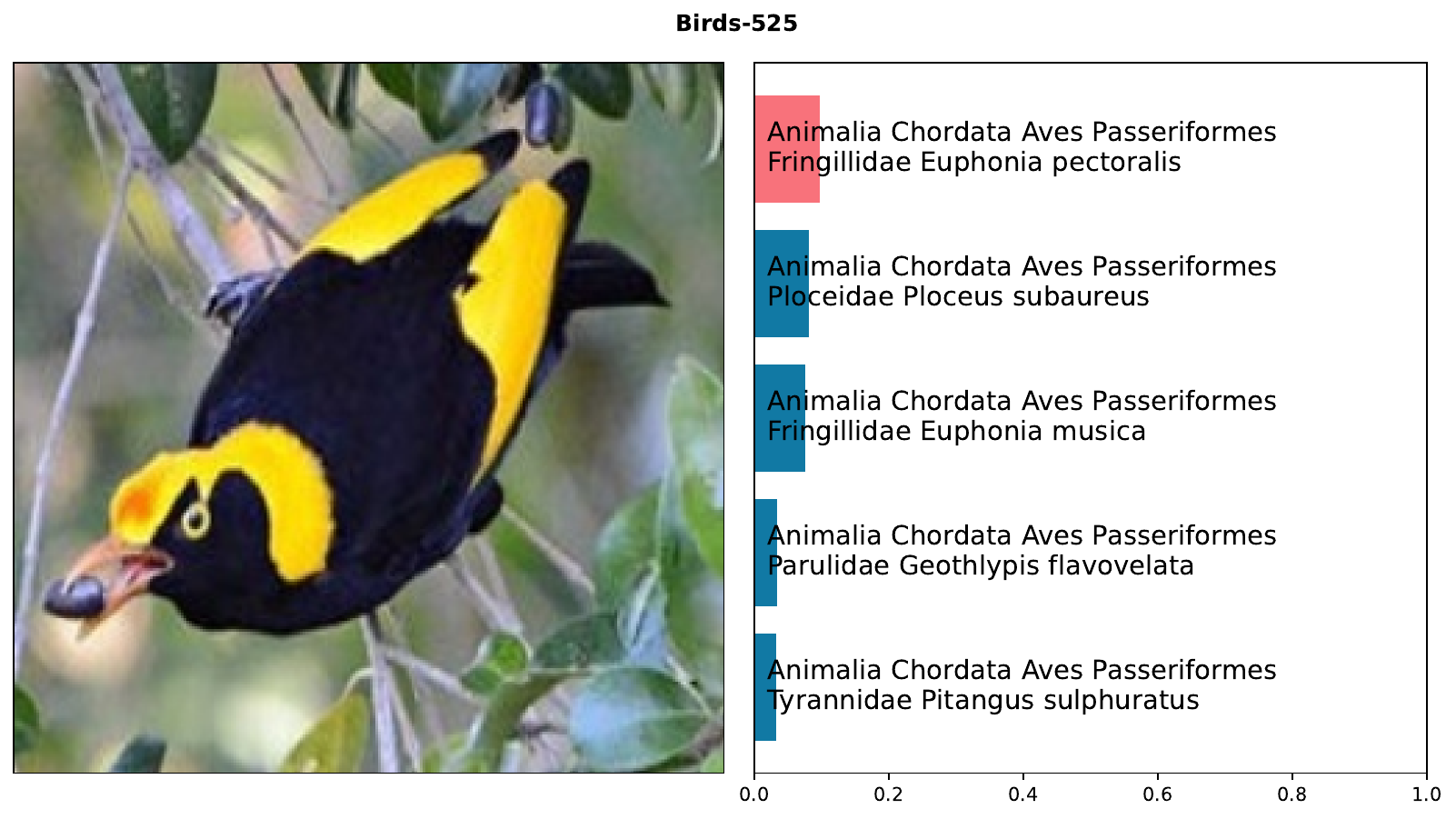}
    \end{subfigure}
    \hfill
    \begin{subfigure}[b]{0.33\textwidth}
        \includegraphics[width=\textwidth]{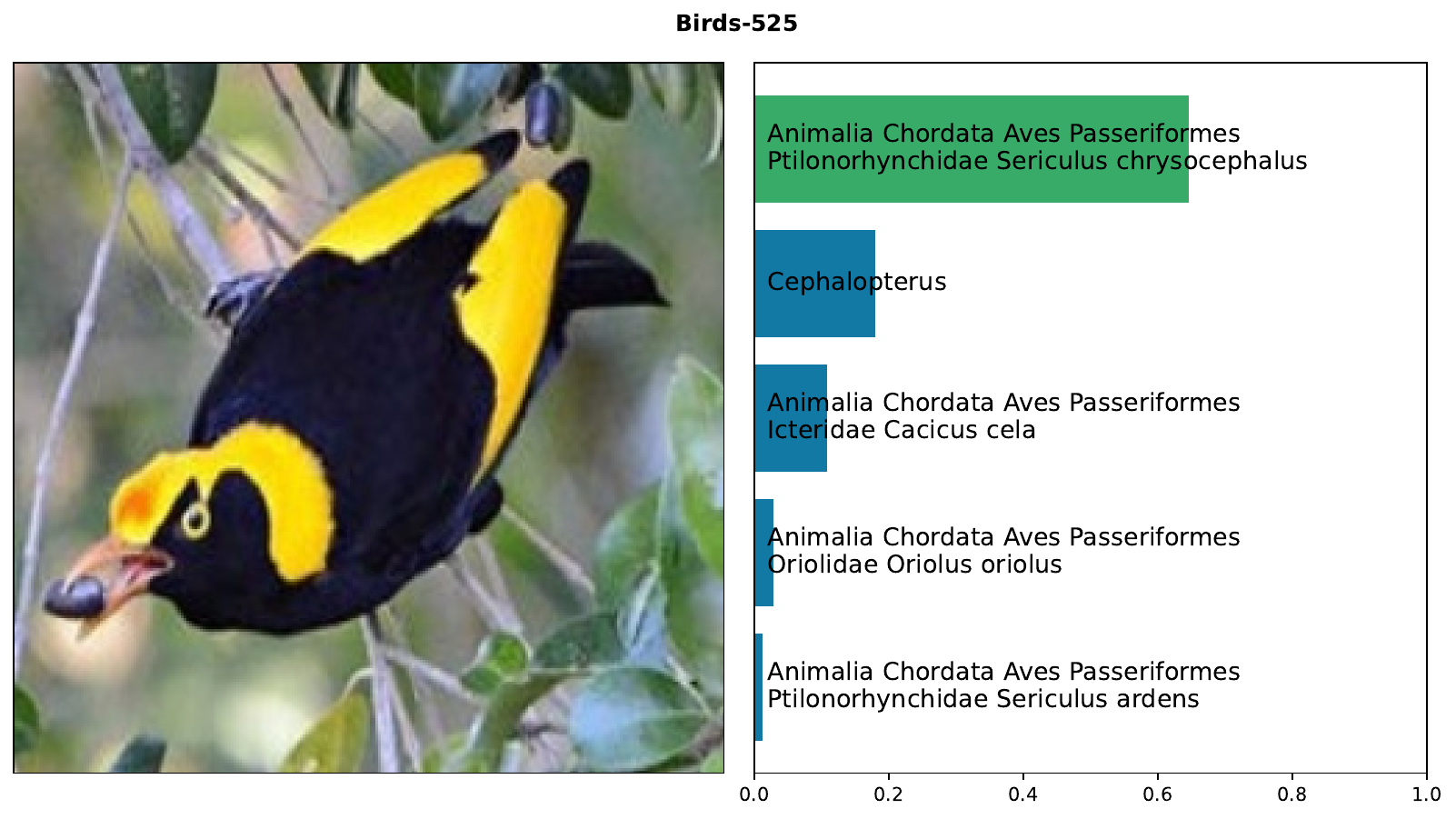}
    \end{subfigure}

    \begin{subfigure}[b]{0.33\textwidth}
        \includegraphics[width=\textwidth]{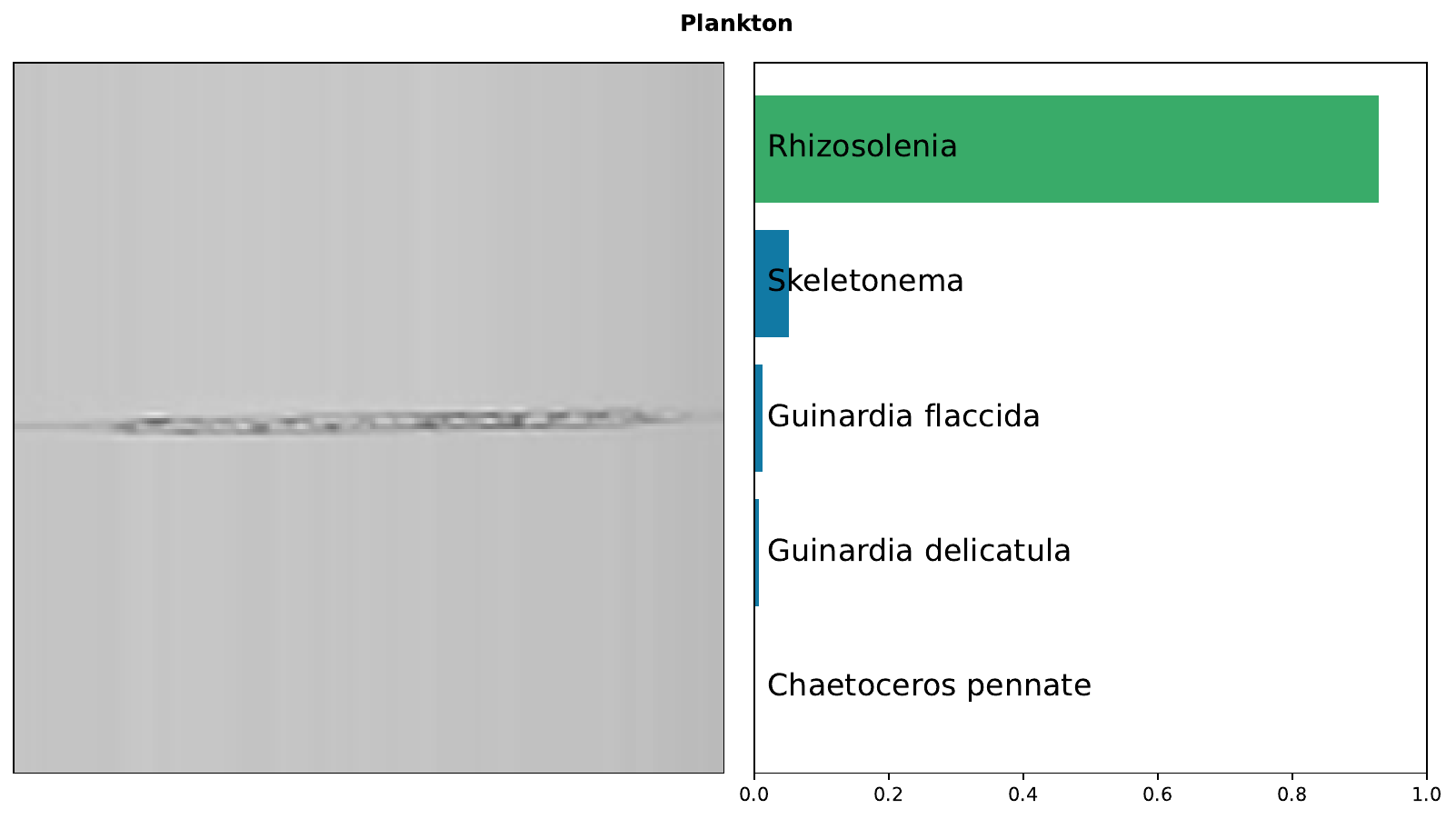}
    \end{subfigure}
    \hfill
    \begin{subfigure}[b]{0.33\textwidth}
        \includegraphics[width=\textwidth]{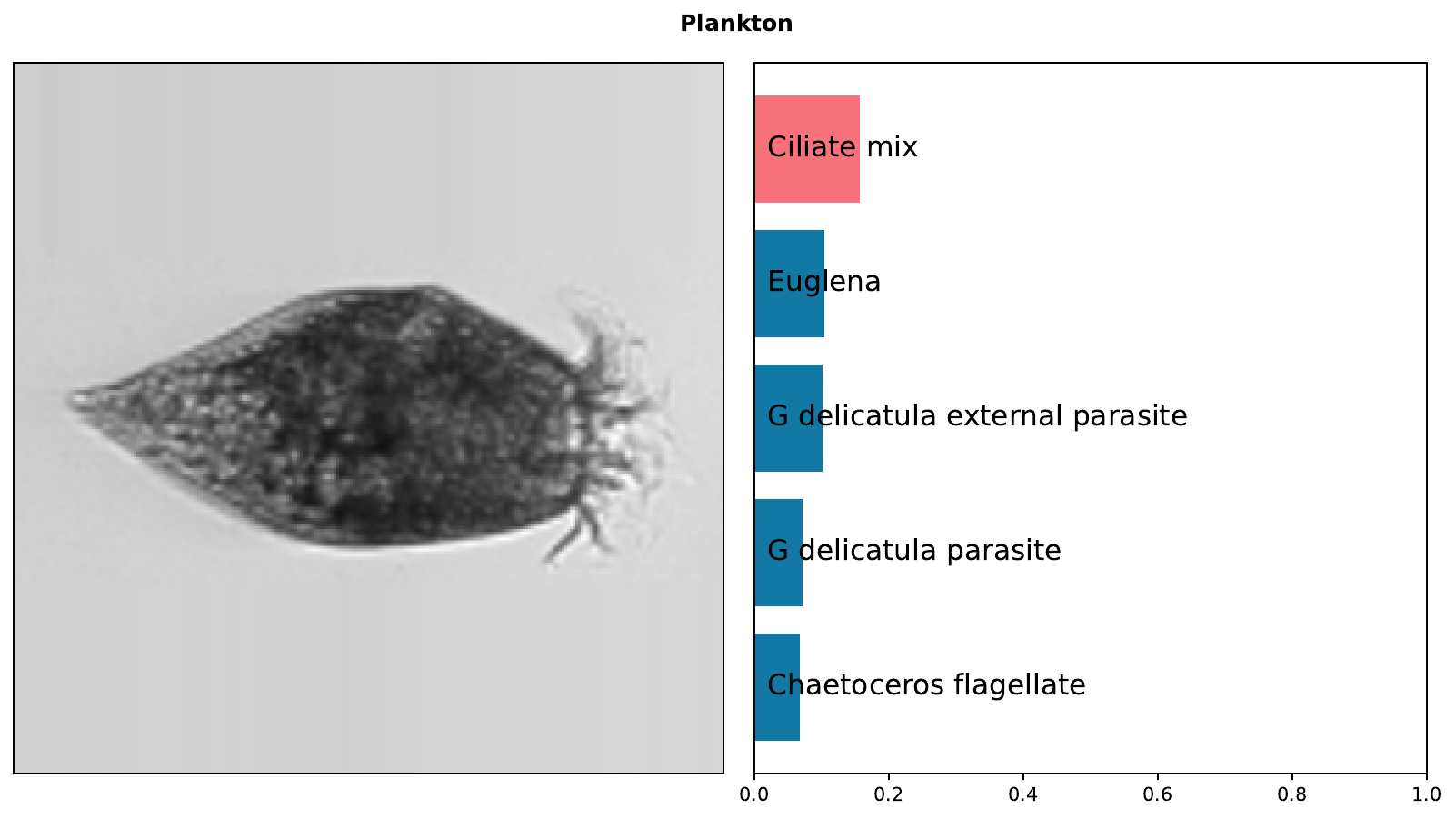}
    \end{subfigure}
    \hfill
    \begin{subfigure}[b]{0.33\textwidth}
        \includegraphics[width=\textwidth]{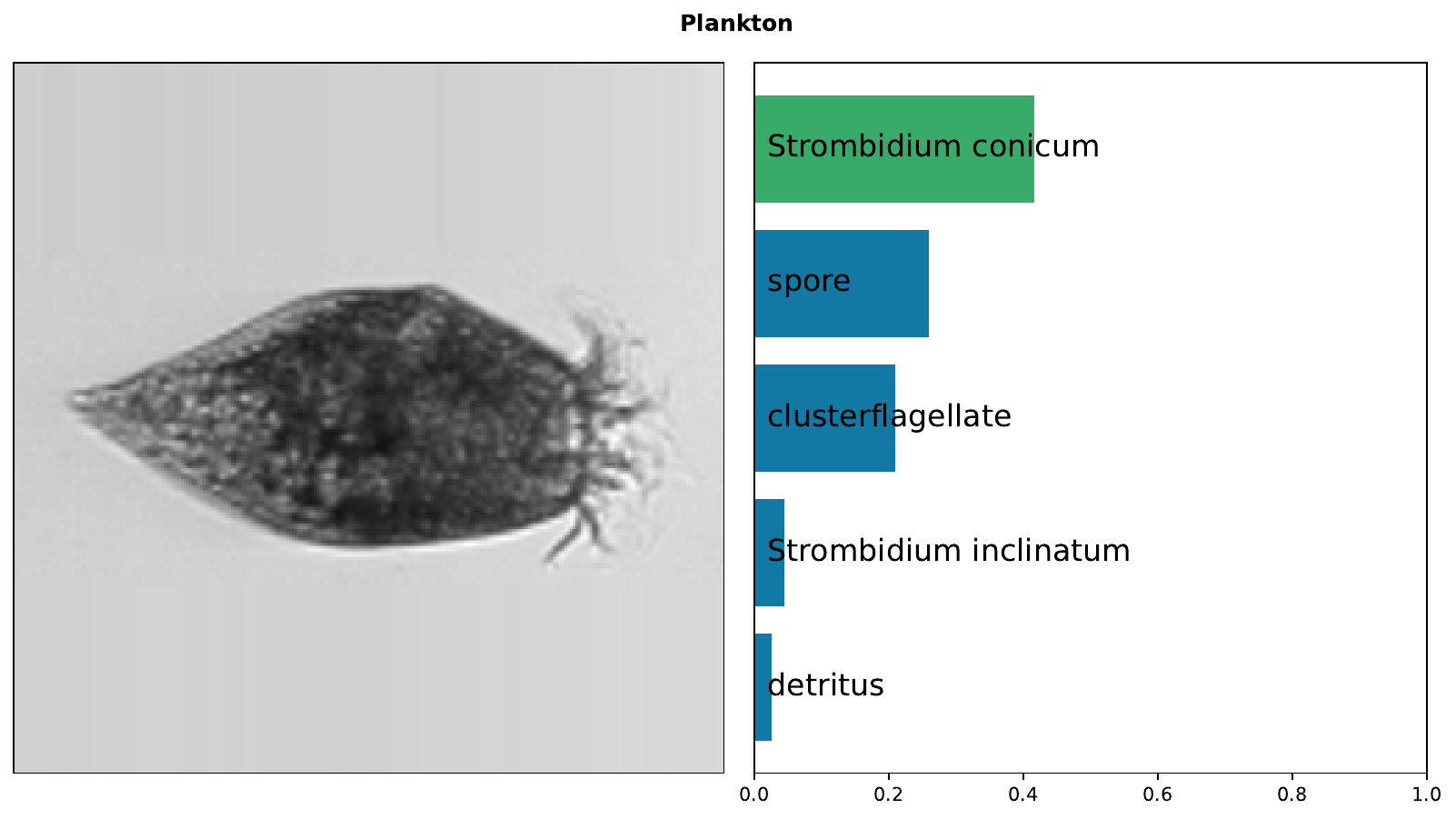}
    \end{subfigure}

    \begin{subfigure}[b]{0.33\textwidth}
        \includegraphics[width=\textwidth]{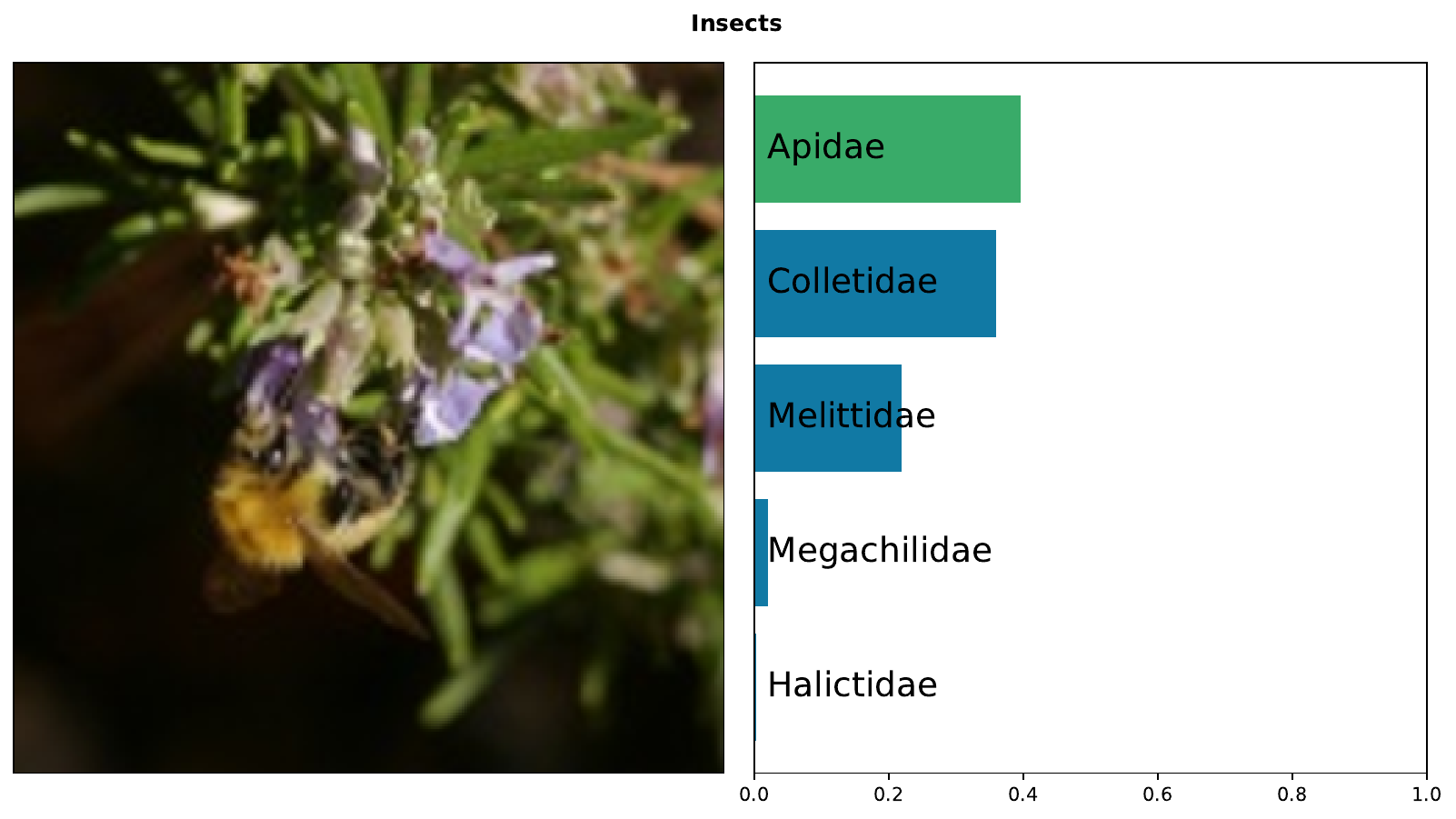}
    \end{subfigure}
    \hfill
    \begin{subfigure}[b]{0.33\textwidth}
        \includegraphics[width=\textwidth]{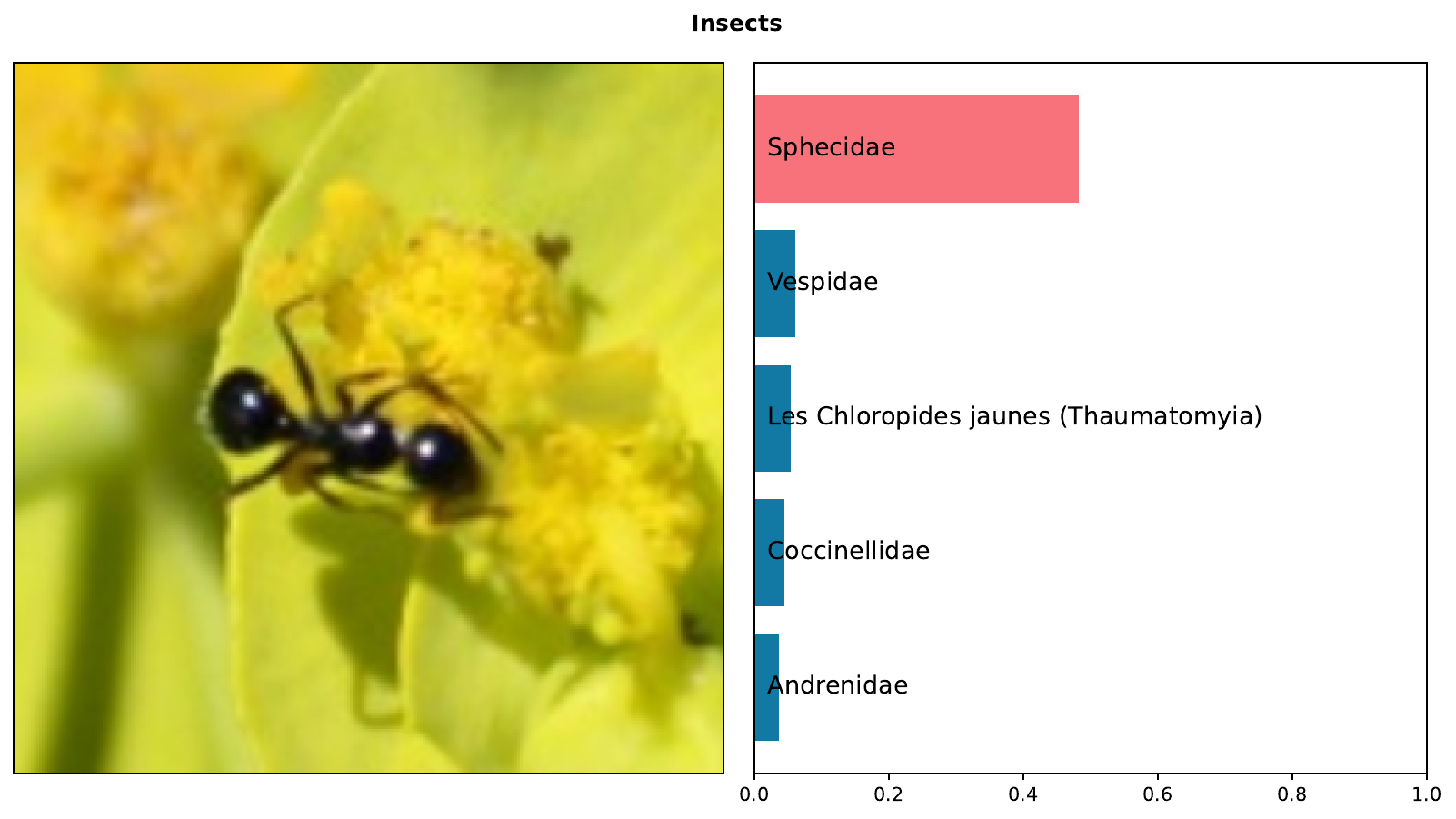}
    \end{subfigure}
    \hfill
    \begin{subfigure}[b]{0.33\textwidth}
        \includegraphics[width=\textwidth]{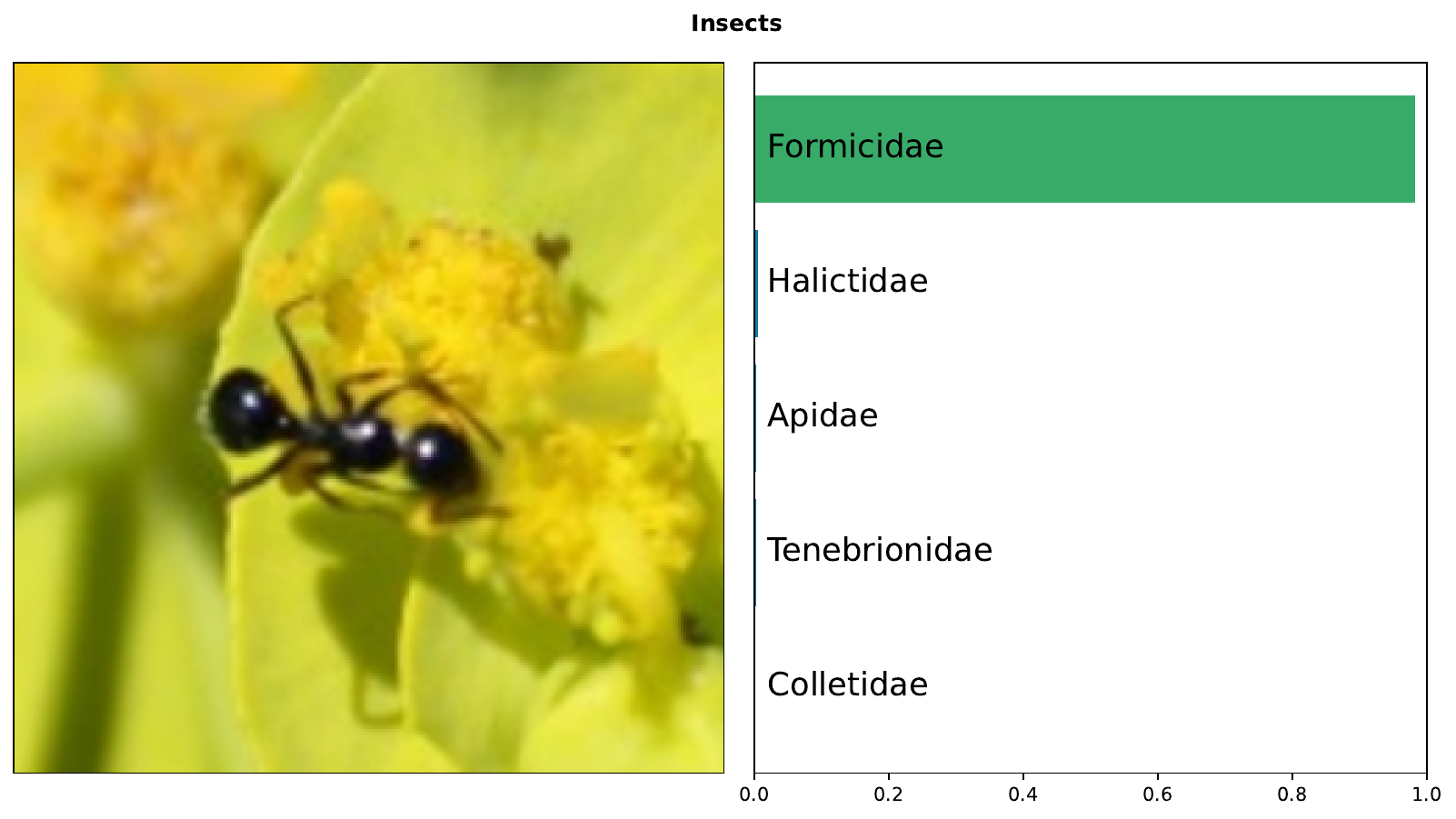}
    \end{subfigure}

    \begin{subfigure}[b]{0.33\textwidth}
        \includegraphics[width=\textwidth]{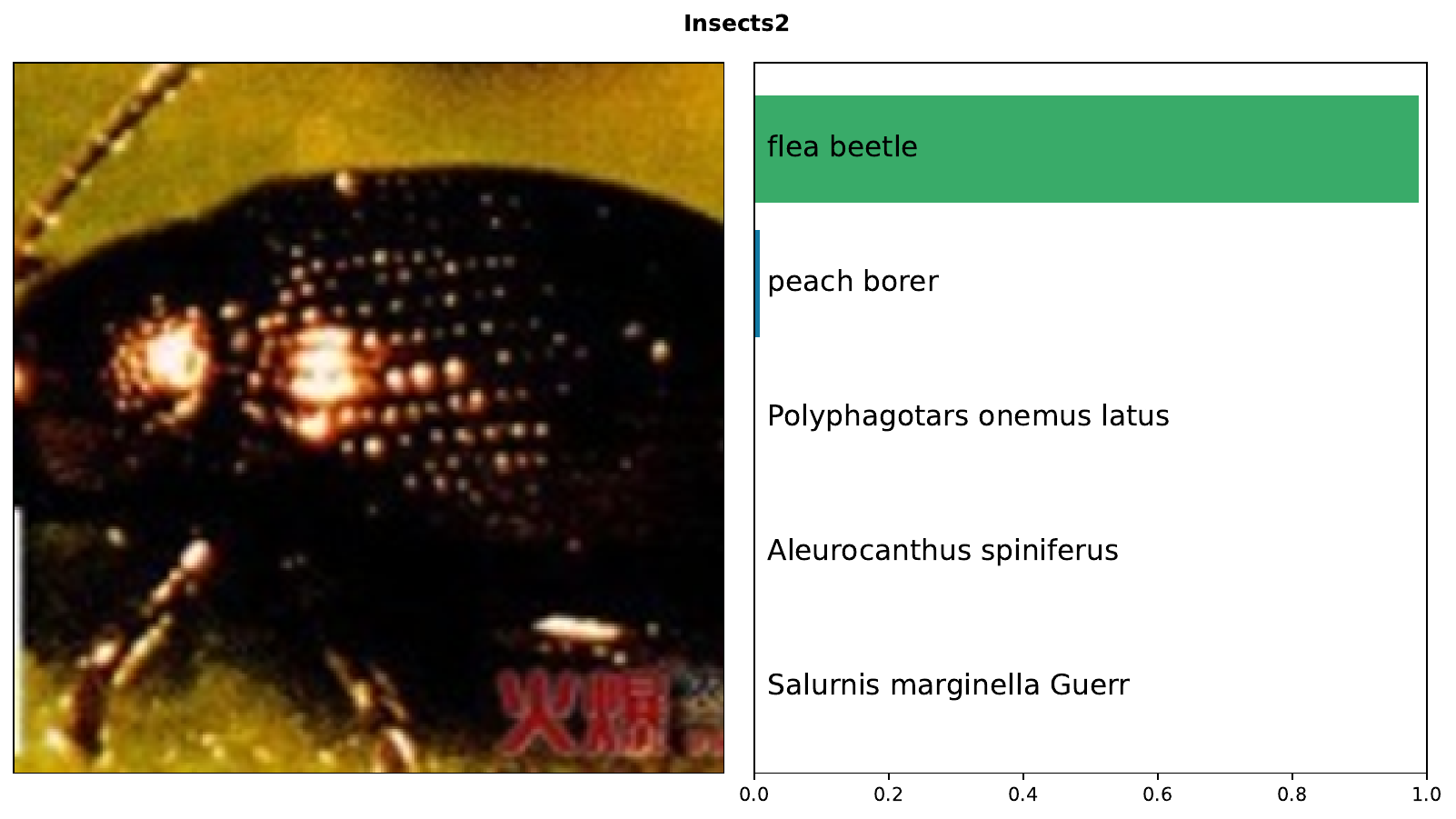}
    \end{subfigure}
    \hfill
    \begin{subfigure}[b]{0.33\textwidth}
        \includegraphics[width=\textwidth]{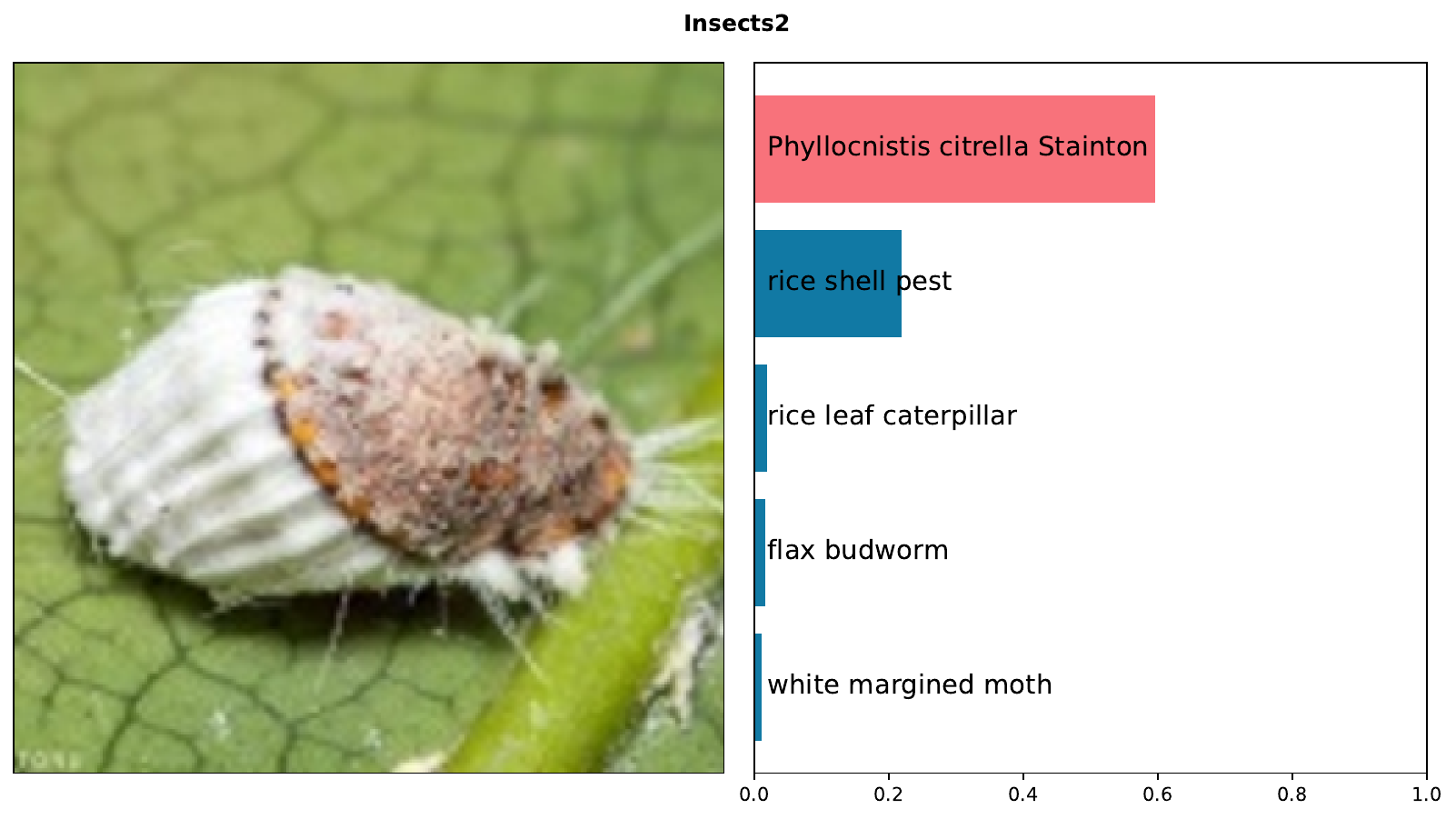}
    \end{subfigure}
    \hfill
    \begin{subfigure}[b]{0.33\textwidth}
        \includegraphics[width=\textwidth]{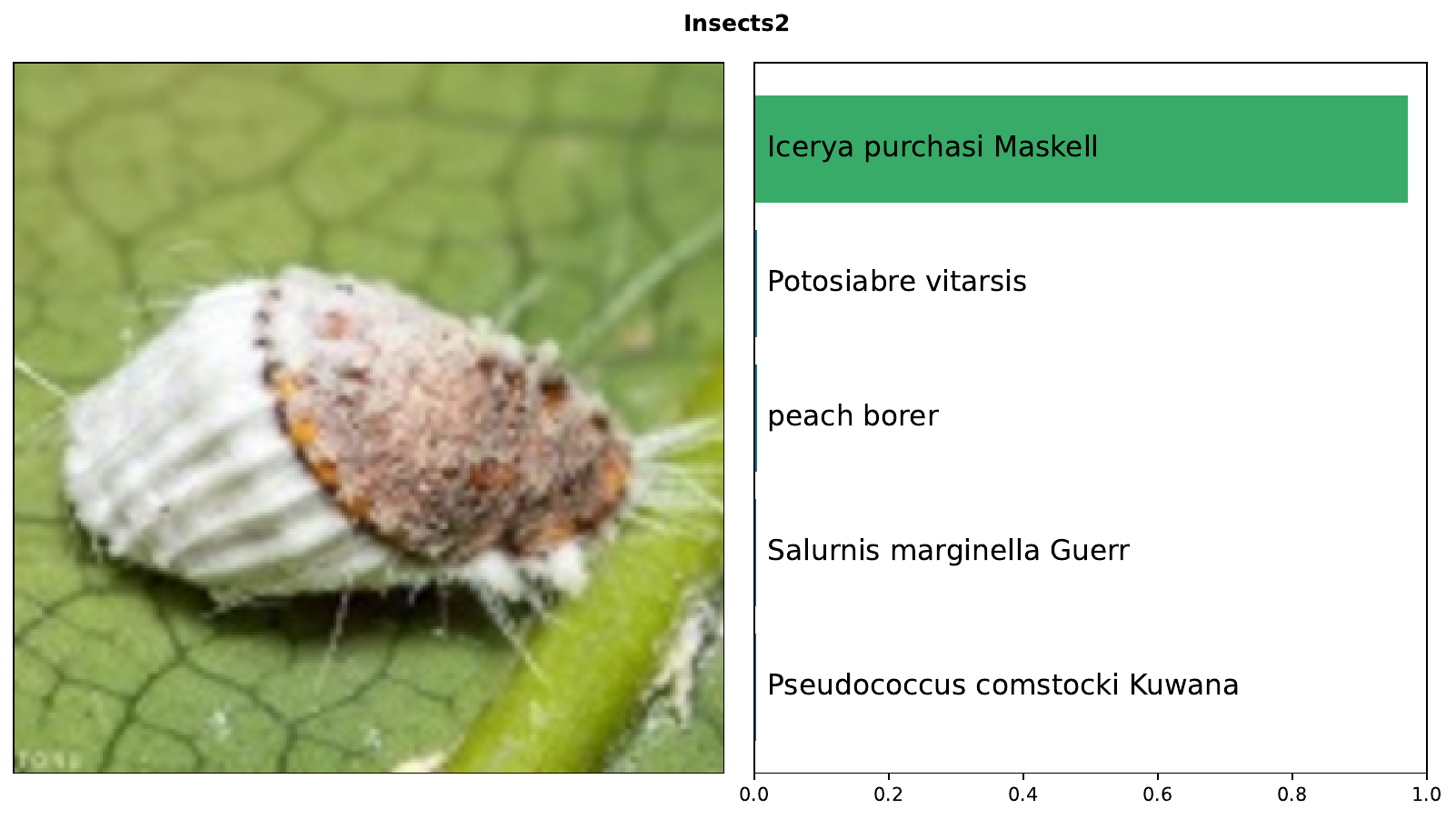}
    \end{subfigure}

    \begin{subfigure}[b]{0.33\textwidth}
        \includegraphics[width=\textwidth]{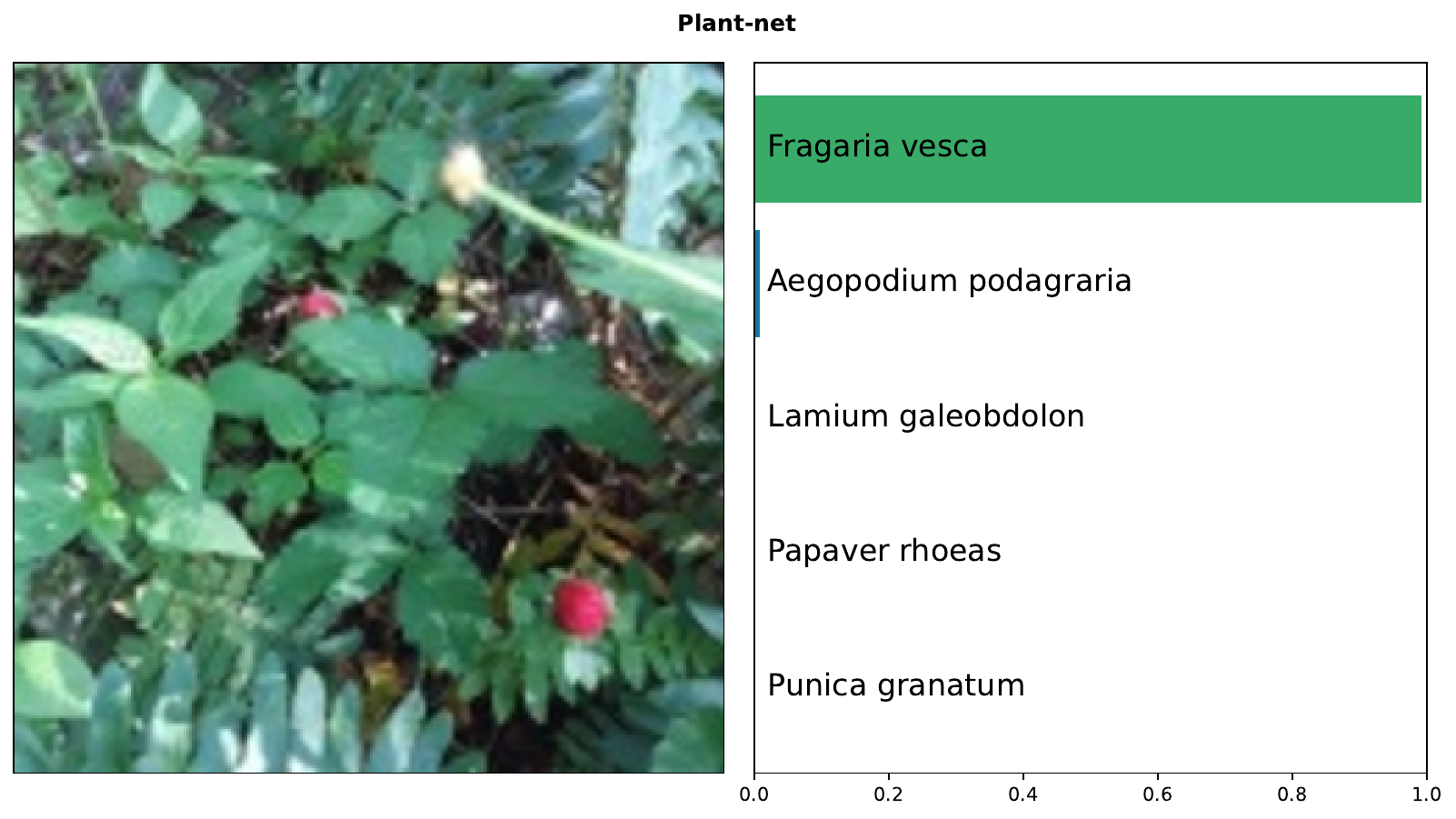}
    \end{subfigure}
    \hfill
    \begin{subfigure}[b]{0.33\textwidth}
        \includegraphics[width=\textwidth]{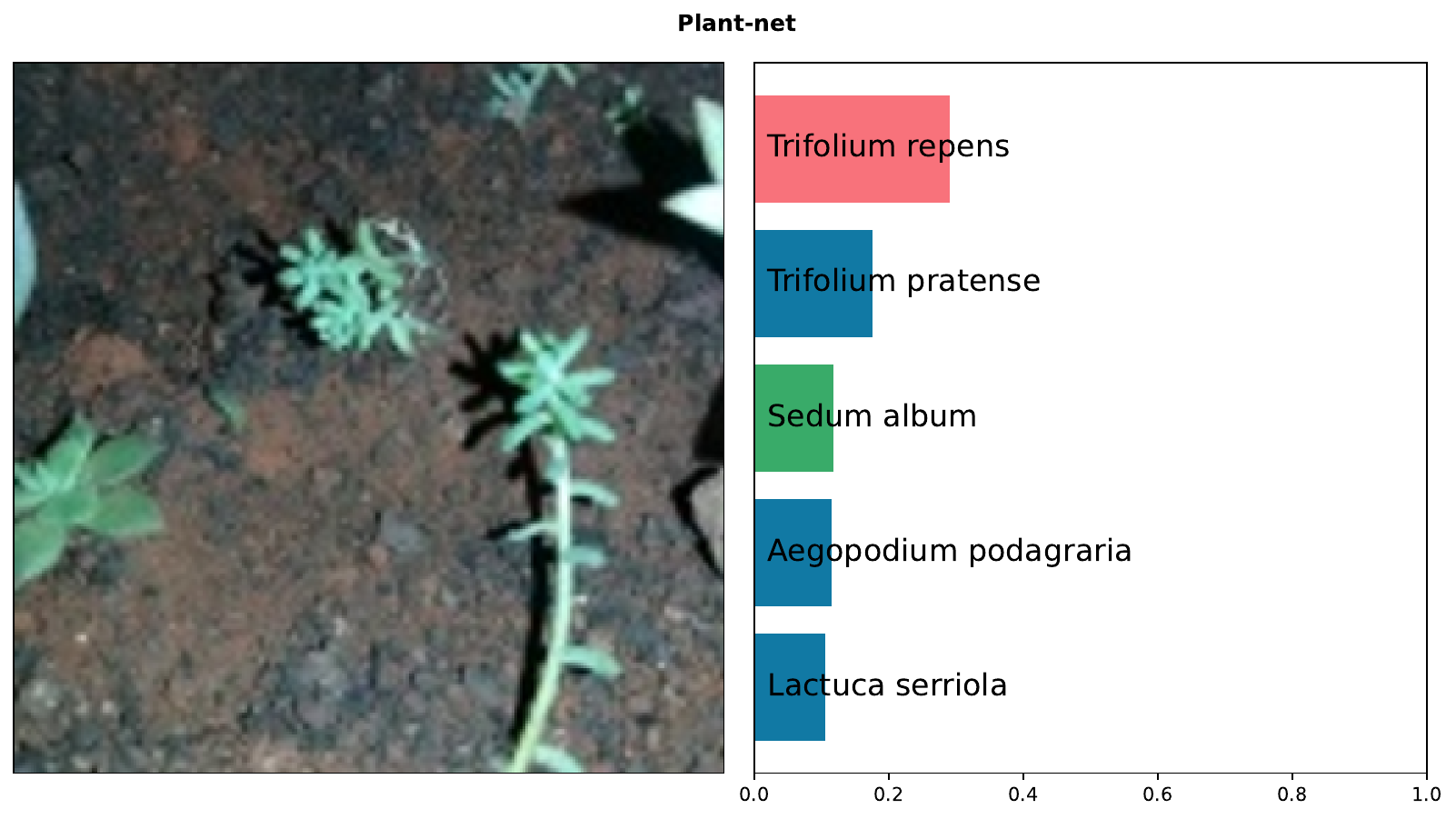}
    \end{subfigure}
    \hfill
    \begin{subfigure}[b]{0.33\textwidth}
        \includegraphics[width=\textwidth]{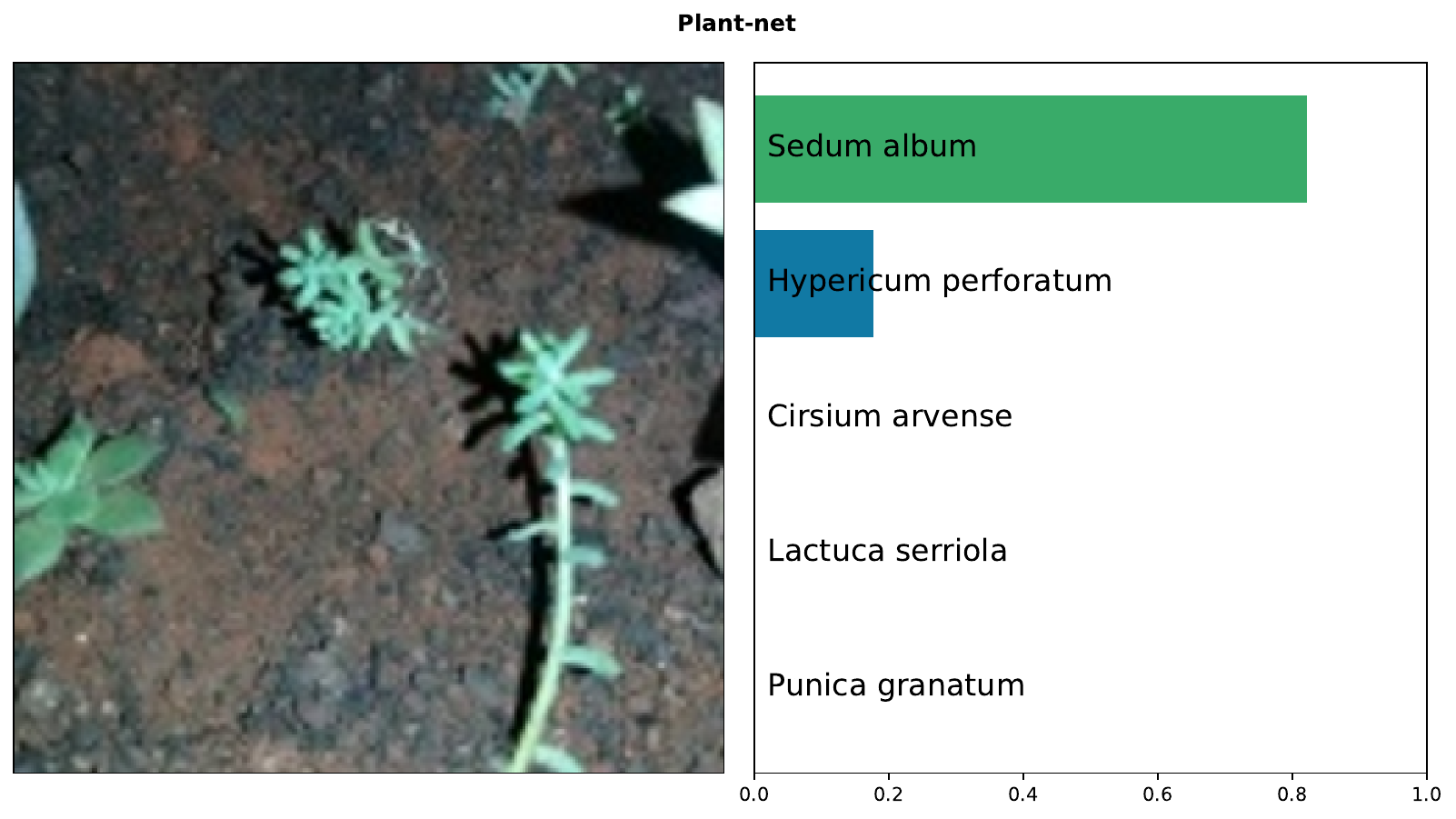}
    \end{subfigure}

    \begin{subfigure}[b]{0.33\textwidth}
        \includegraphics[width=\textwidth]{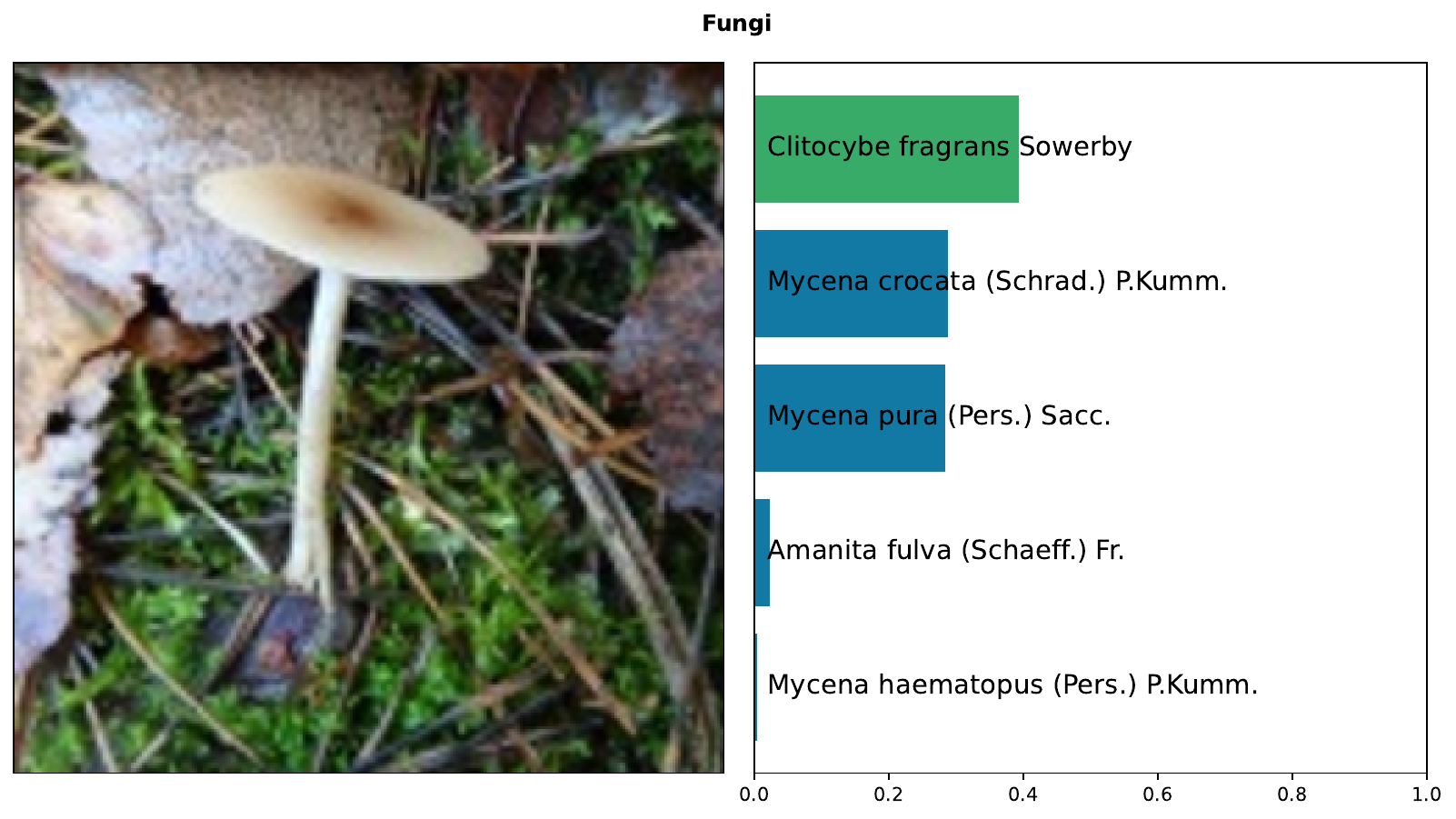}
    \end{subfigure}
    \hfill
    \begin{subfigure}[b]{0.33\textwidth}
        \includegraphics[width=\textwidth]{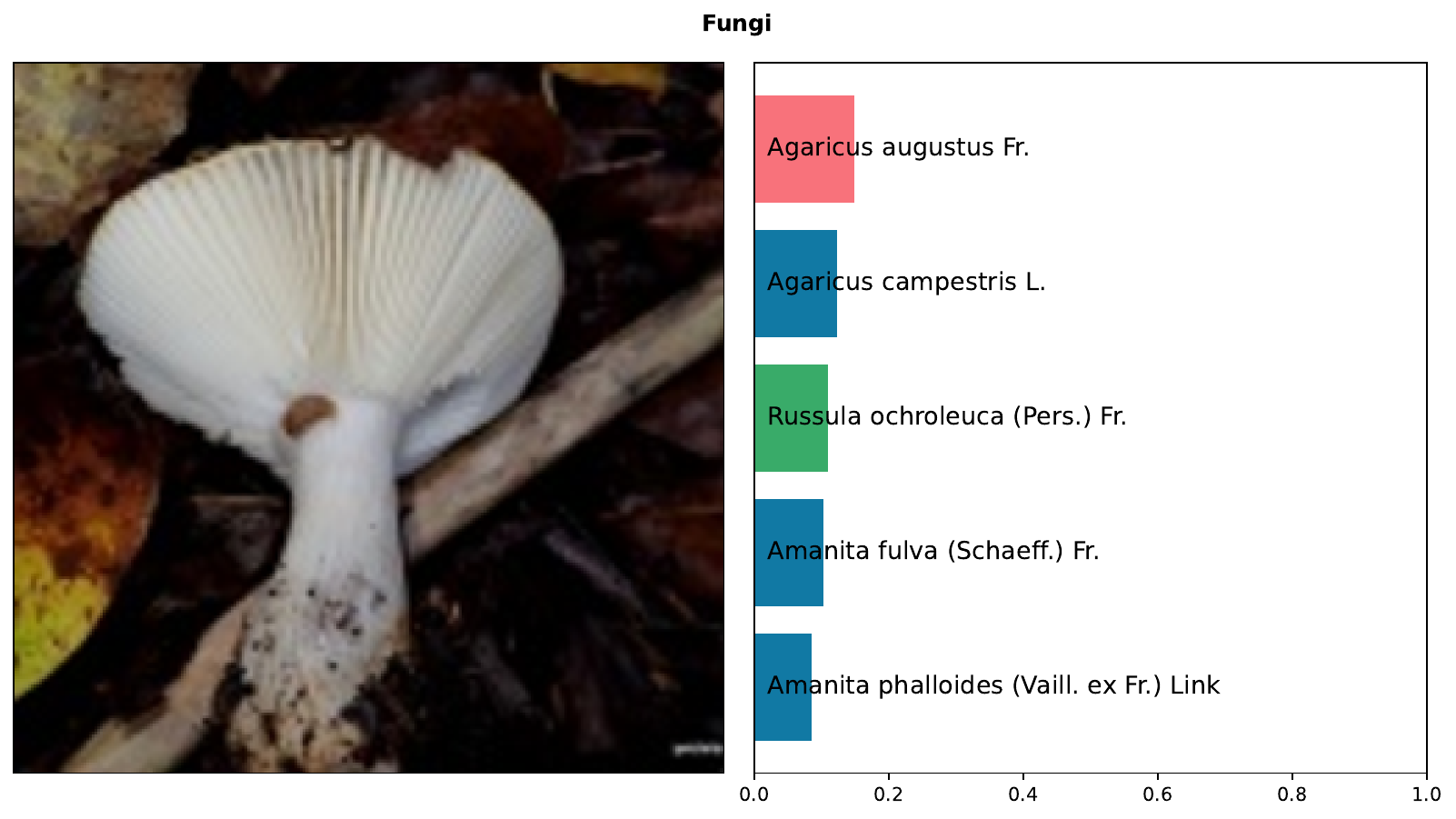}
    \end{subfigure}
    \hfill
    \begin{subfigure}[b]{0.33\textwidth}
        \includegraphics[width=\textwidth]{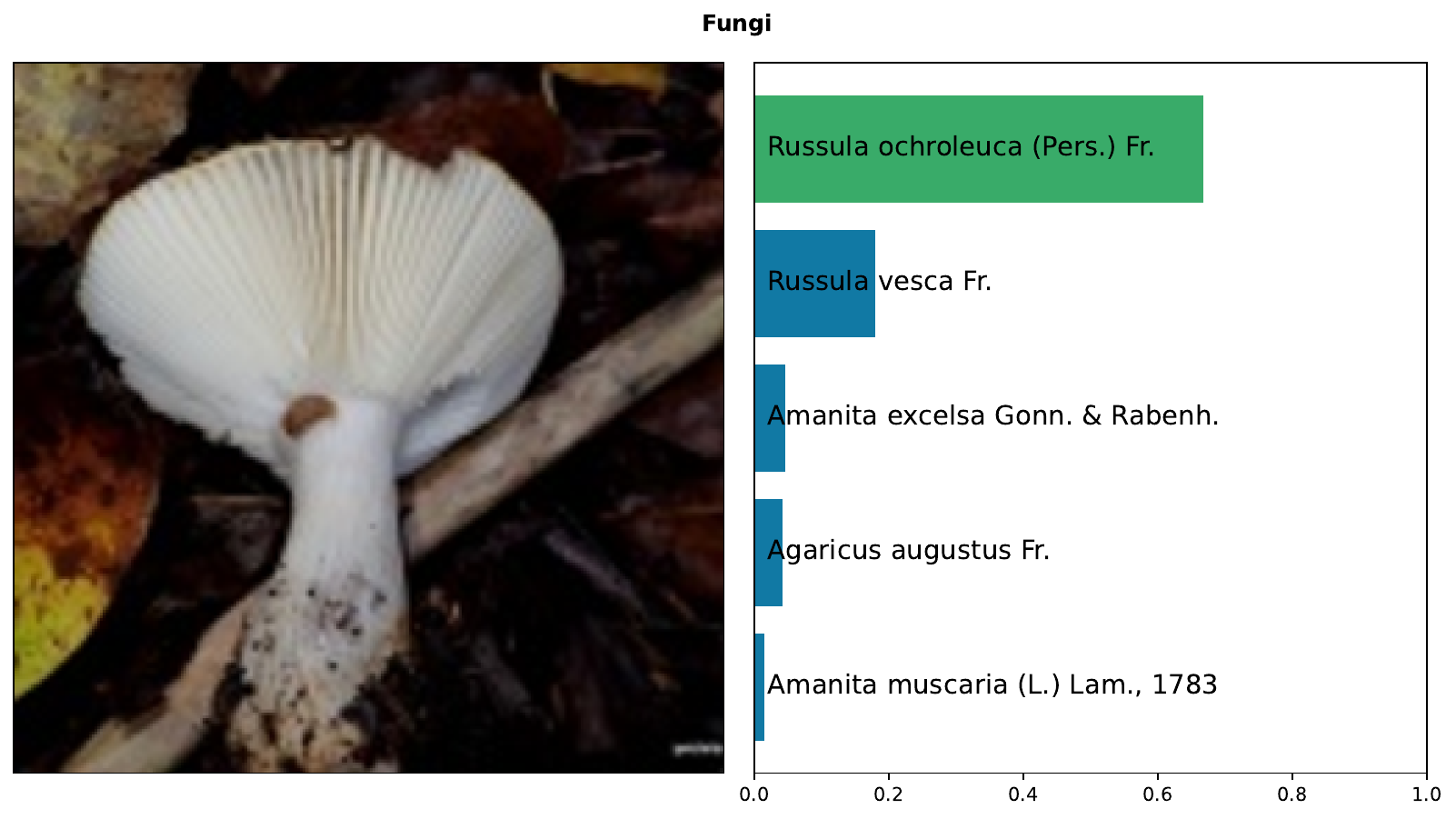}
    \end{subfigure}
    \vskip -6pt
    \caption{
    Example predictions for \modelname{} and CLIP on Birds 525, Plankton, Insects, Insects2, PlantNet and Fungi tasks.
    Ground truth labels are green; incorrect predictions are red.
    Left: Correct \modelname{} predictions.
    Center, Right: Images that CLIP incorrectly labels, but \modelname{} correctly labels.
    }
    \label{fig:example-predictions1}
    \vskip -8pt
\end{figure*}

\begin{figure*}[t]
    \begin{subfigure}[b]{0.33\textwidth}
        \includegraphics[width=\textwidth]{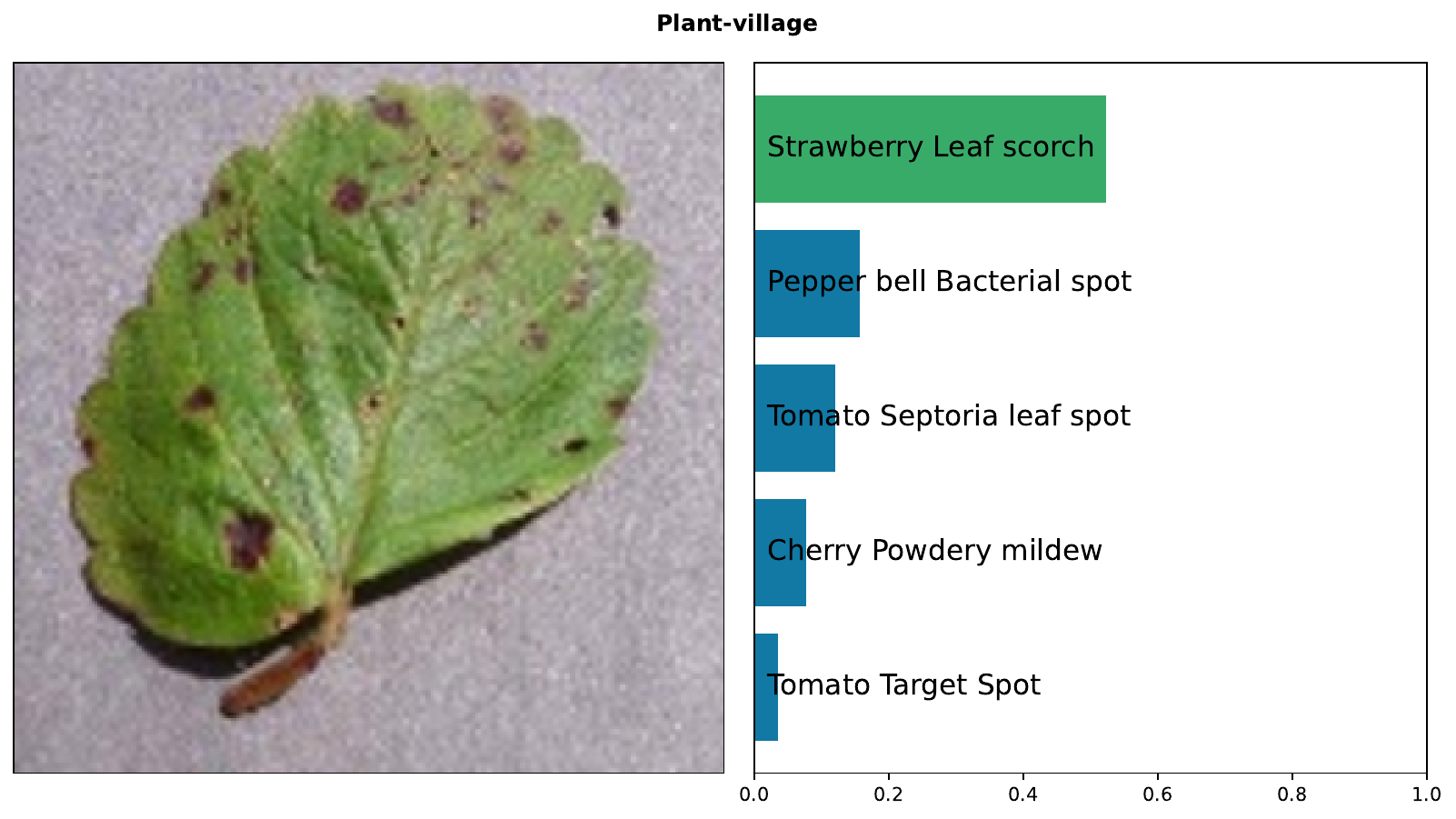}
    \end{subfigure}
    \hfill
    \begin{subfigure}[b]{0.33\textwidth}
        \includegraphics[width=\textwidth]{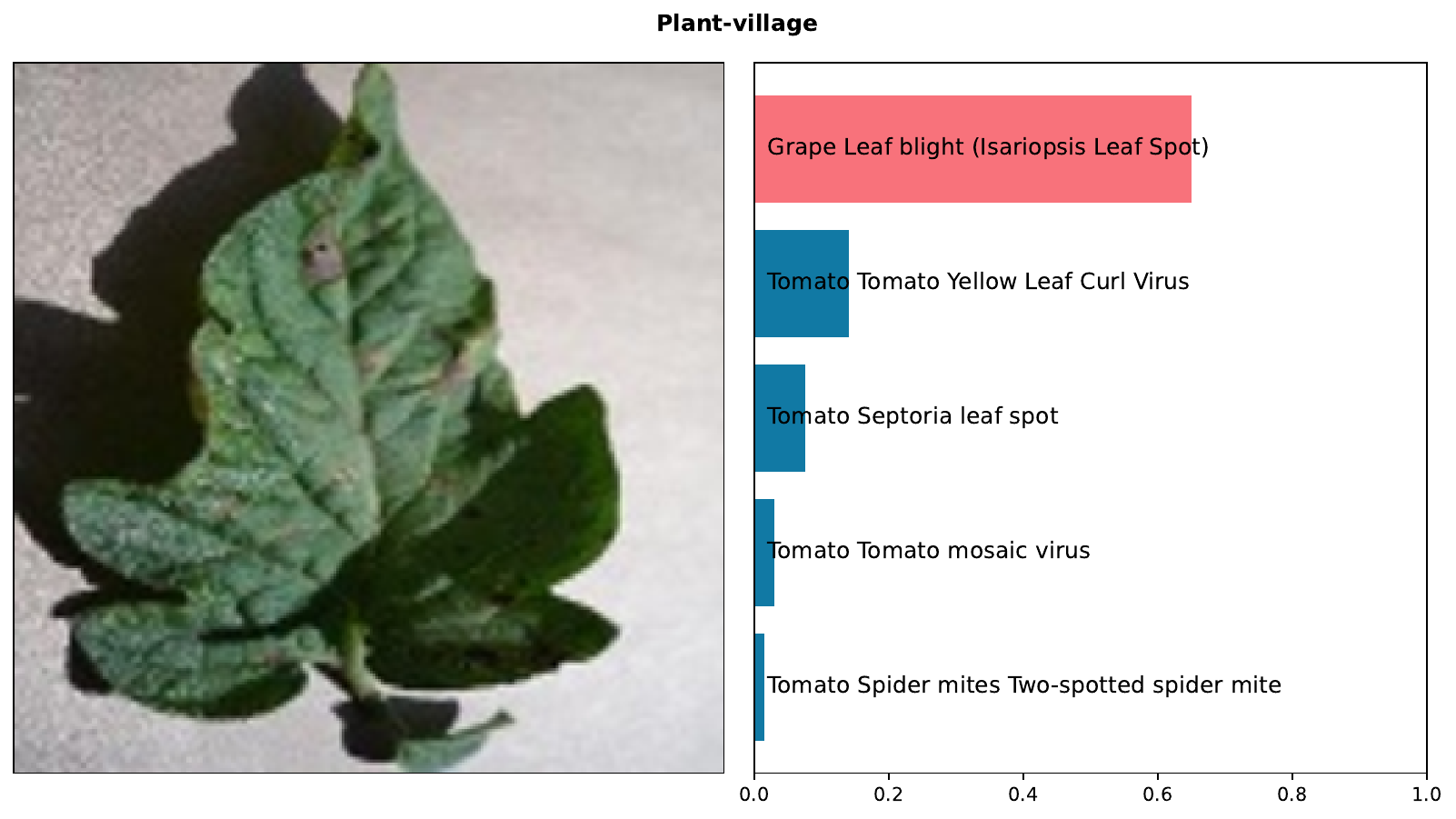}
    \end{subfigure}
    \hfill
    \begin{subfigure}[b]{0.33\textwidth}
        \includegraphics[width=\textwidth]{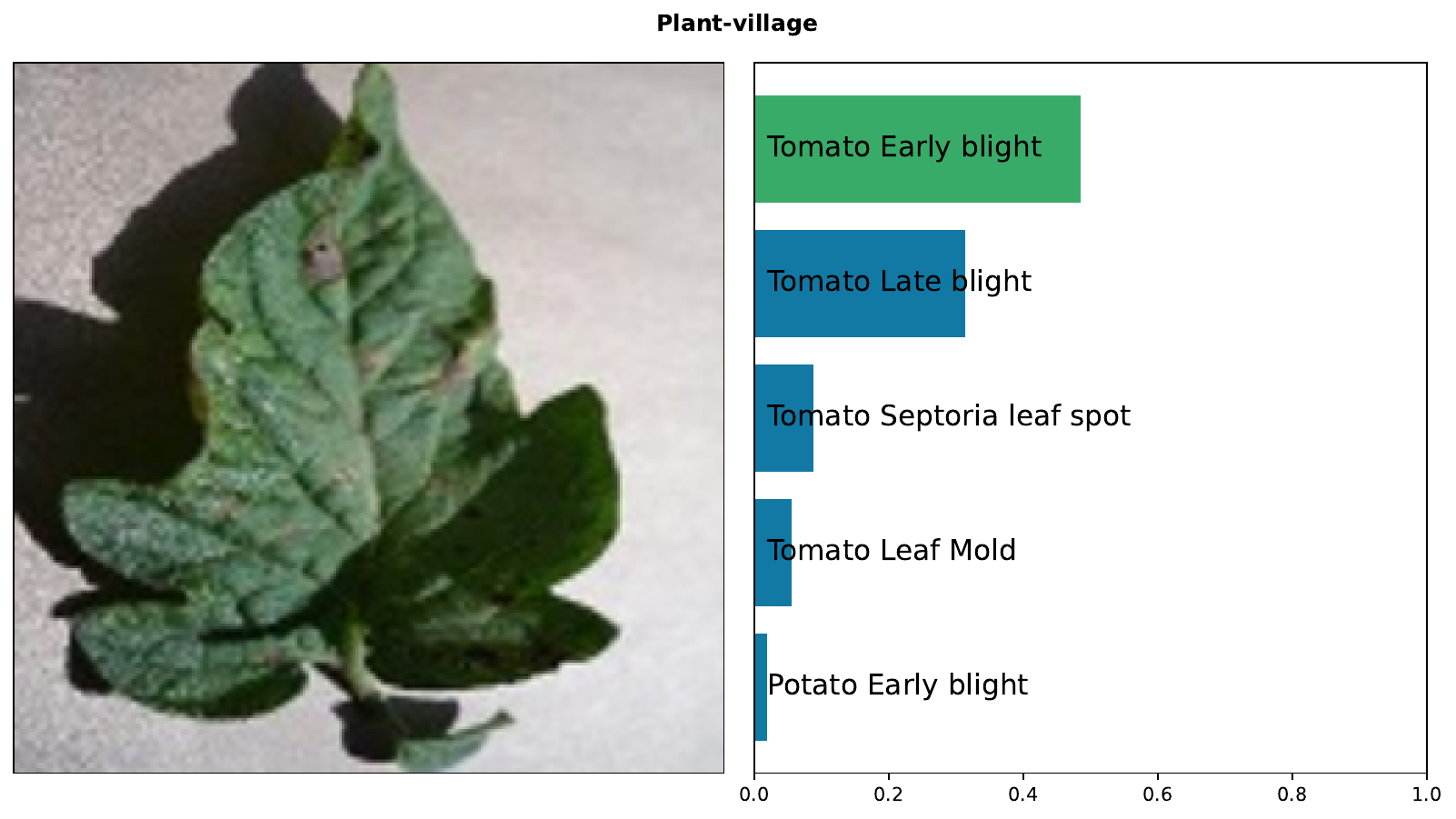}
    \end{subfigure}

    \begin{subfigure}[b]{0.33\textwidth}
        \includegraphics[width=\textwidth]{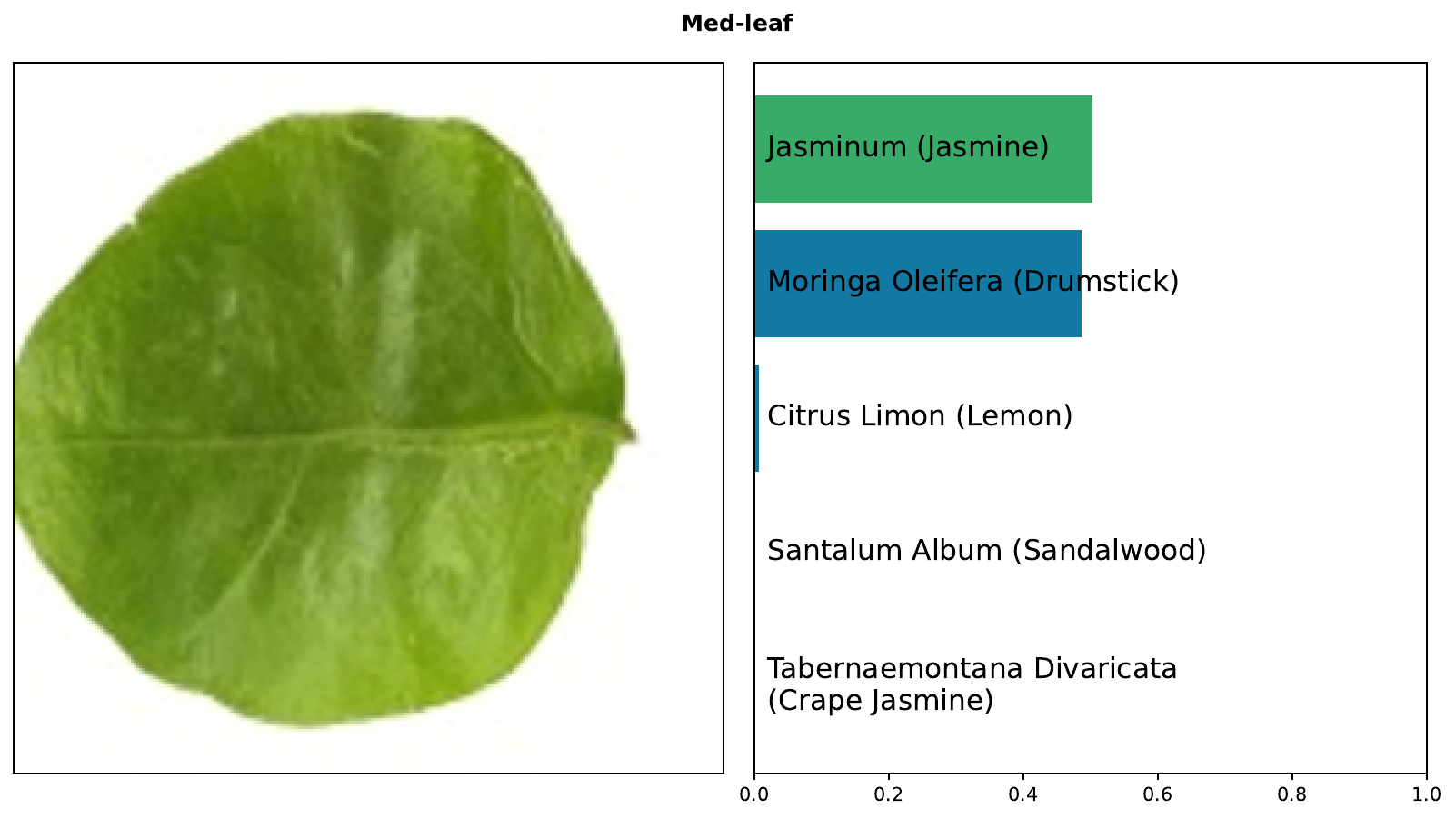}
    \end{subfigure}
    \hfill
    \begin{subfigure}[b]{0.33\textwidth}
        \includegraphics[width=\textwidth]{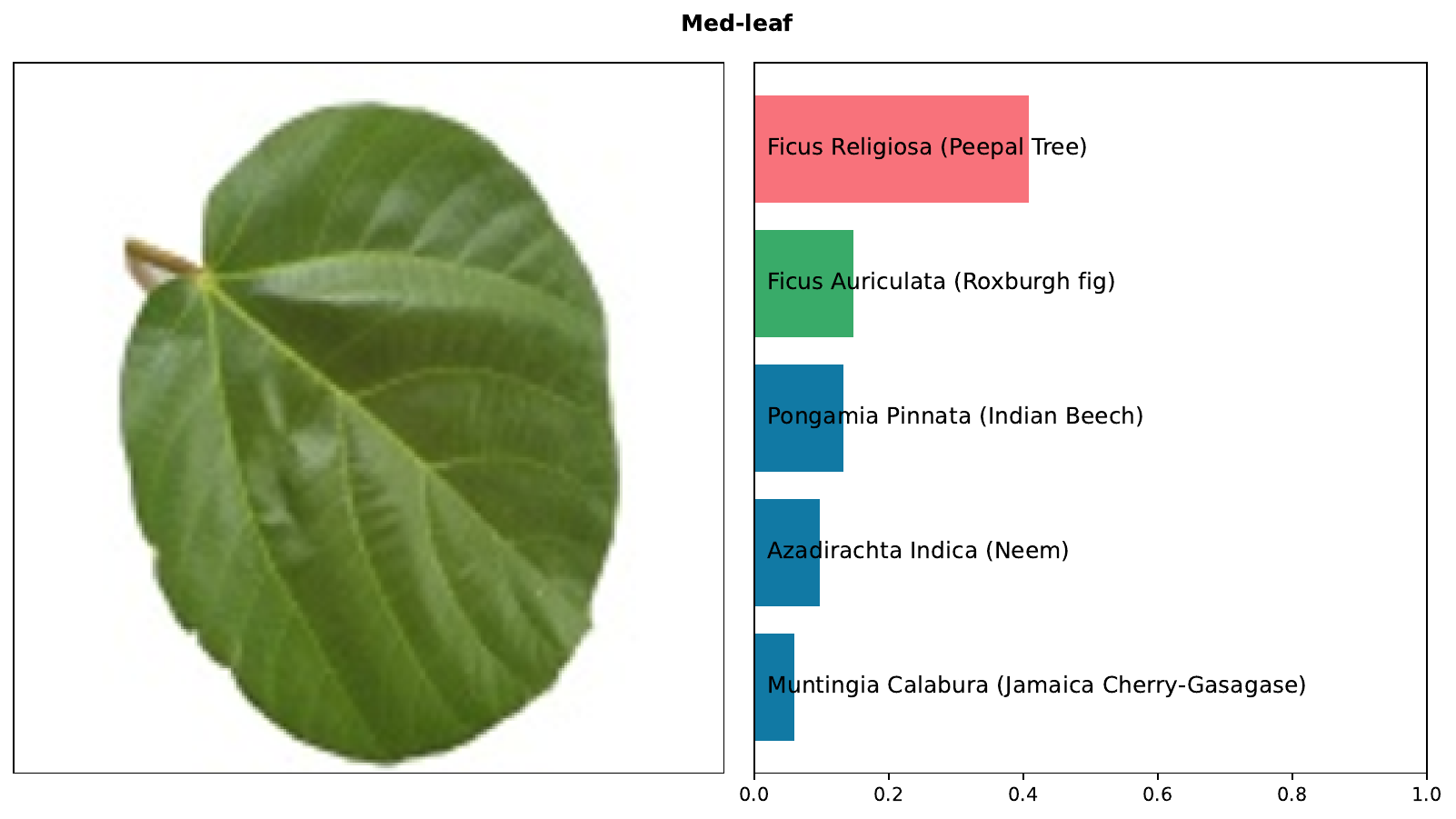}
    \end{subfigure}
    \hfill
    \begin{subfigure}[b]{0.33\textwidth}
        \includegraphics[width=\textwidth]{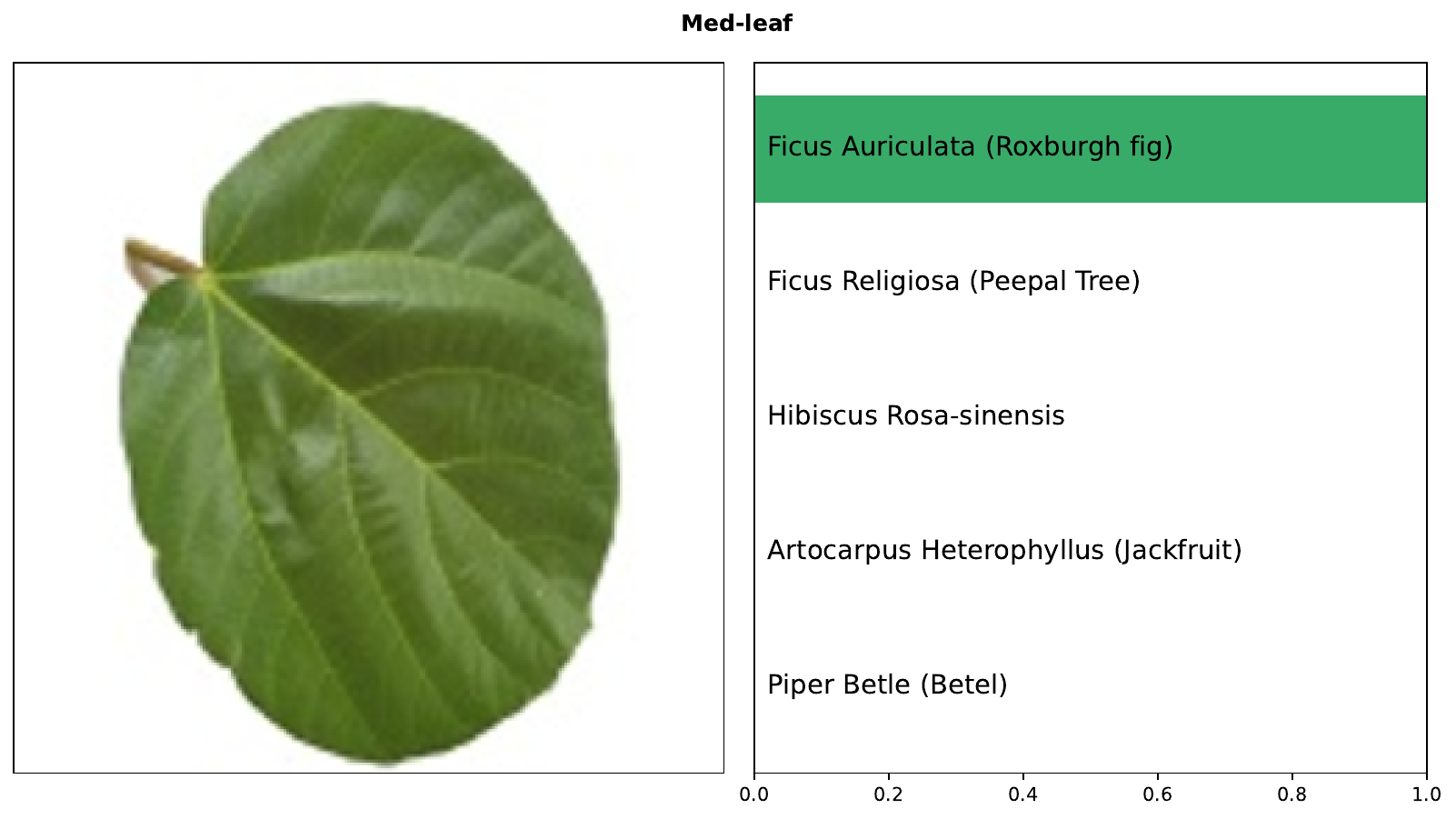}
    \end{subfigure}

    \begin{subfigure}[b]{0.33\textwidth}
        \includegraphics[width=\textwidth]{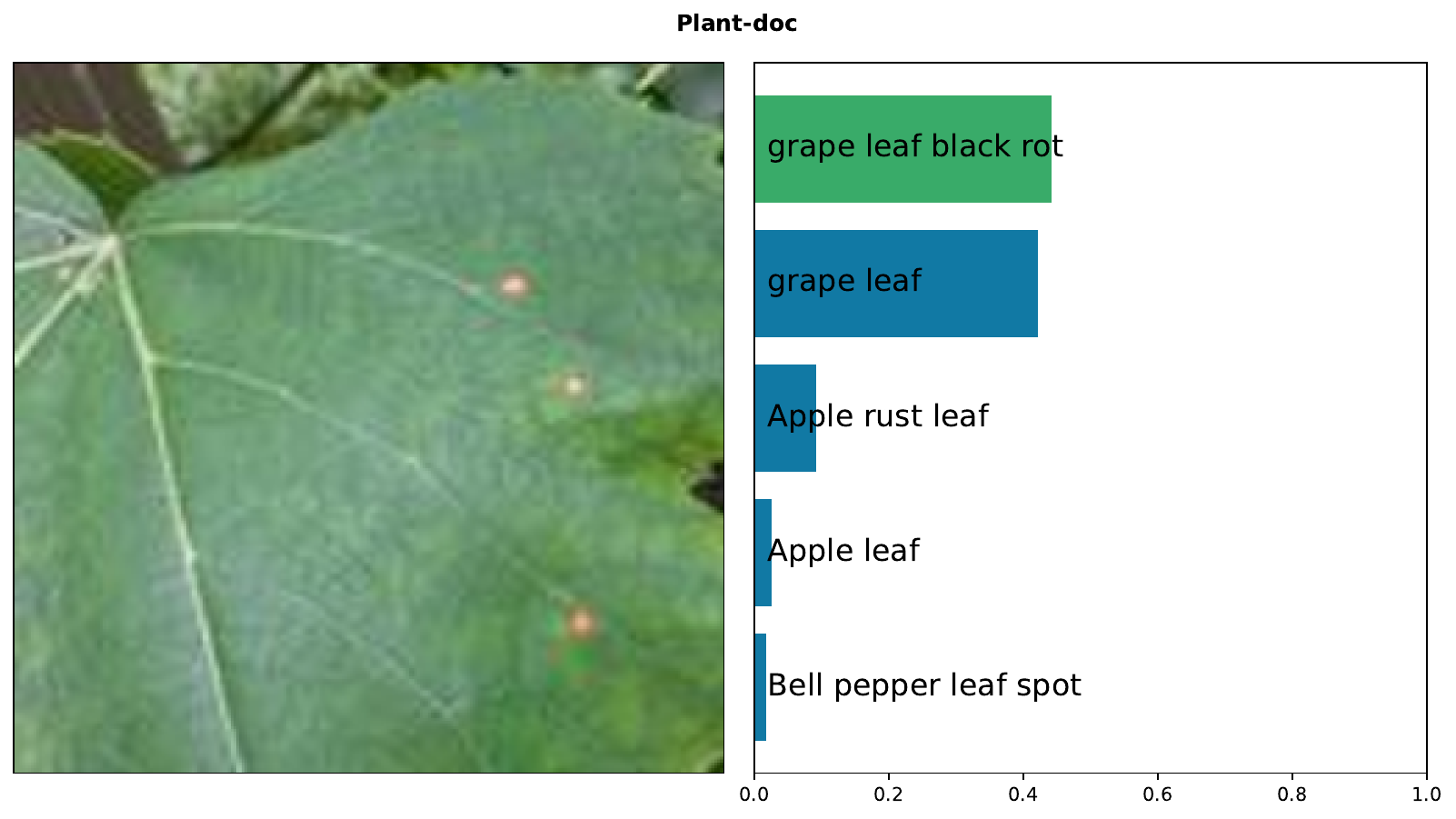}
    \end{subfigure}
    \hfill
    \begin{subfigure}[b]{0.33\textwidth}
        \includegraphics[width=\textwidth]{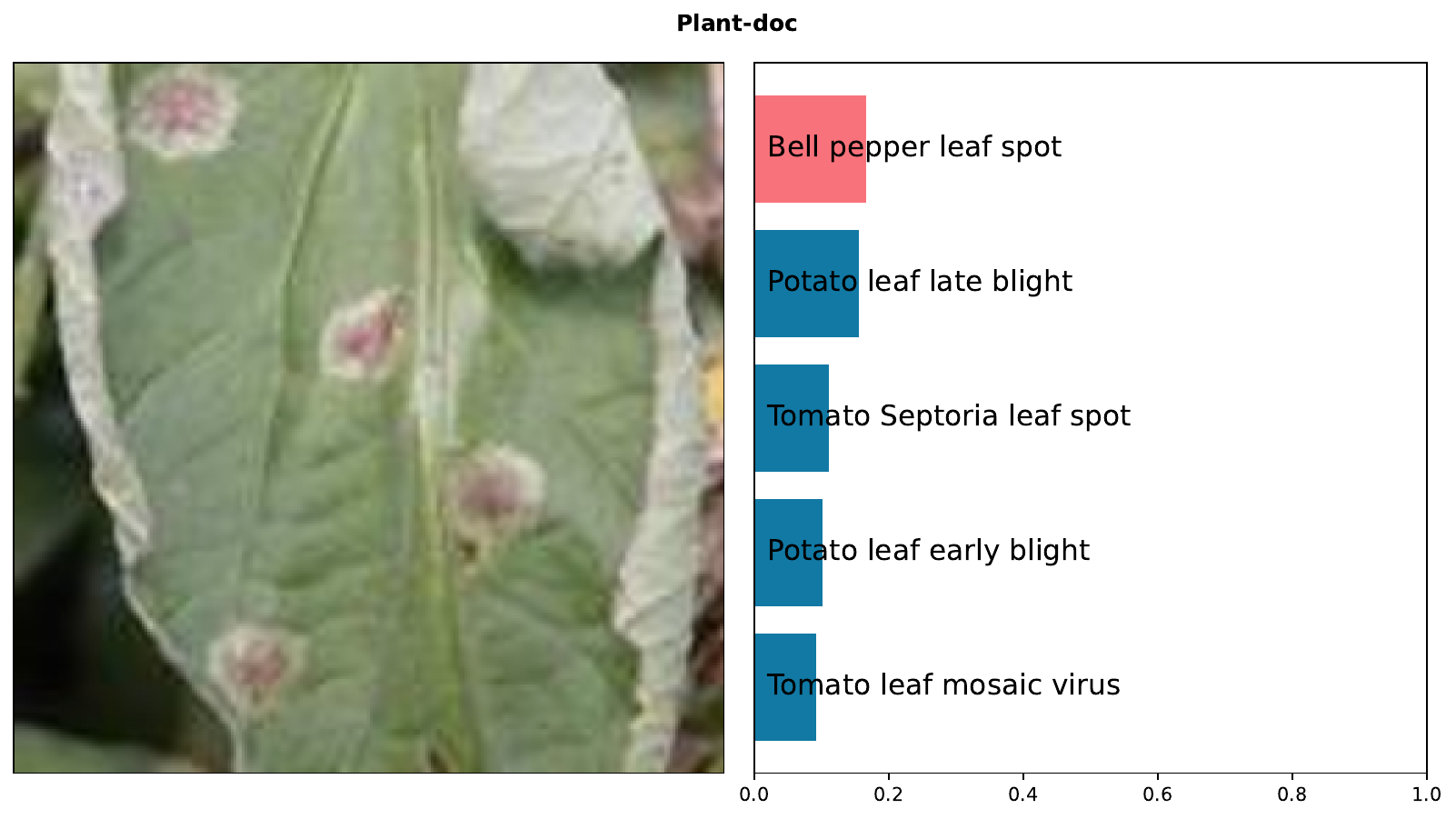}
    \end{subfigure}
    \hfill
    \begin{subfigure}[b]{0.33\textwidth}
        \includegraphics[width=\textwidth]{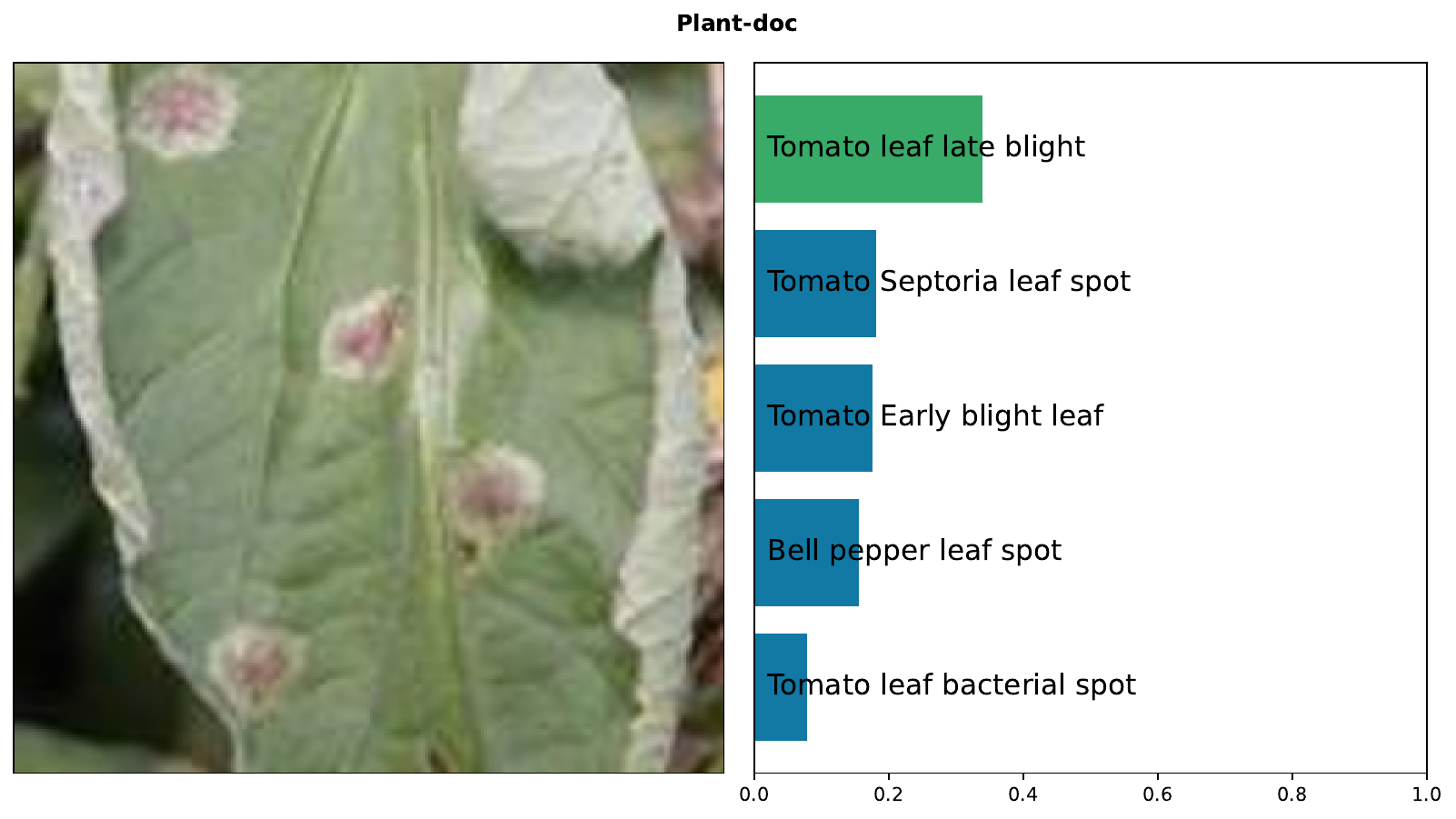}
    \end{subfigure}

    \begin{subfigure}[b]{0.33\textwidth}
        \includegraphics[width=\textwidth]{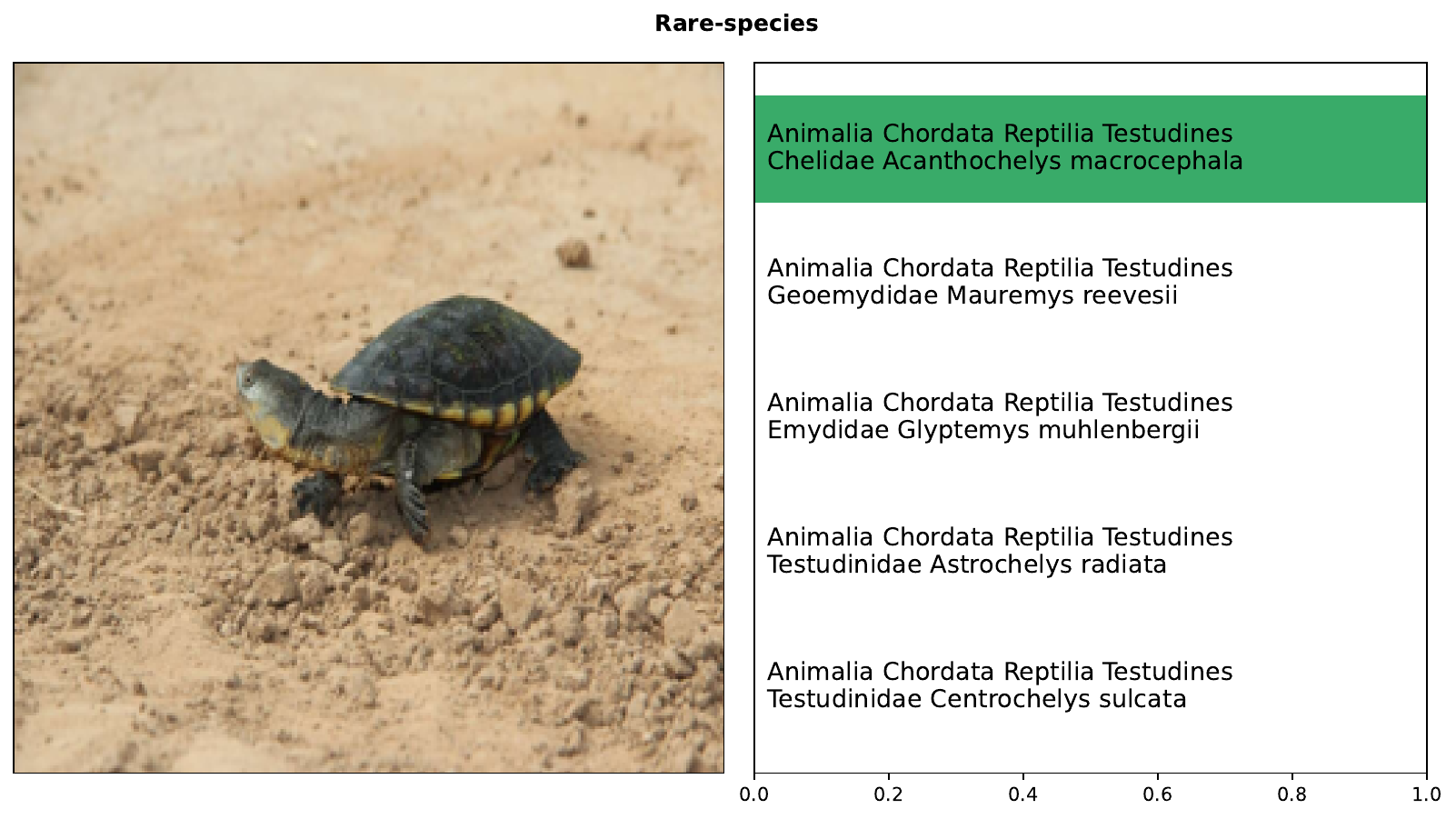}
    \end{subfigure}
    \hfill
    \begin{subfigure}[b]{0.33\textwidth}
        \includegraphics[width=\textwidth]{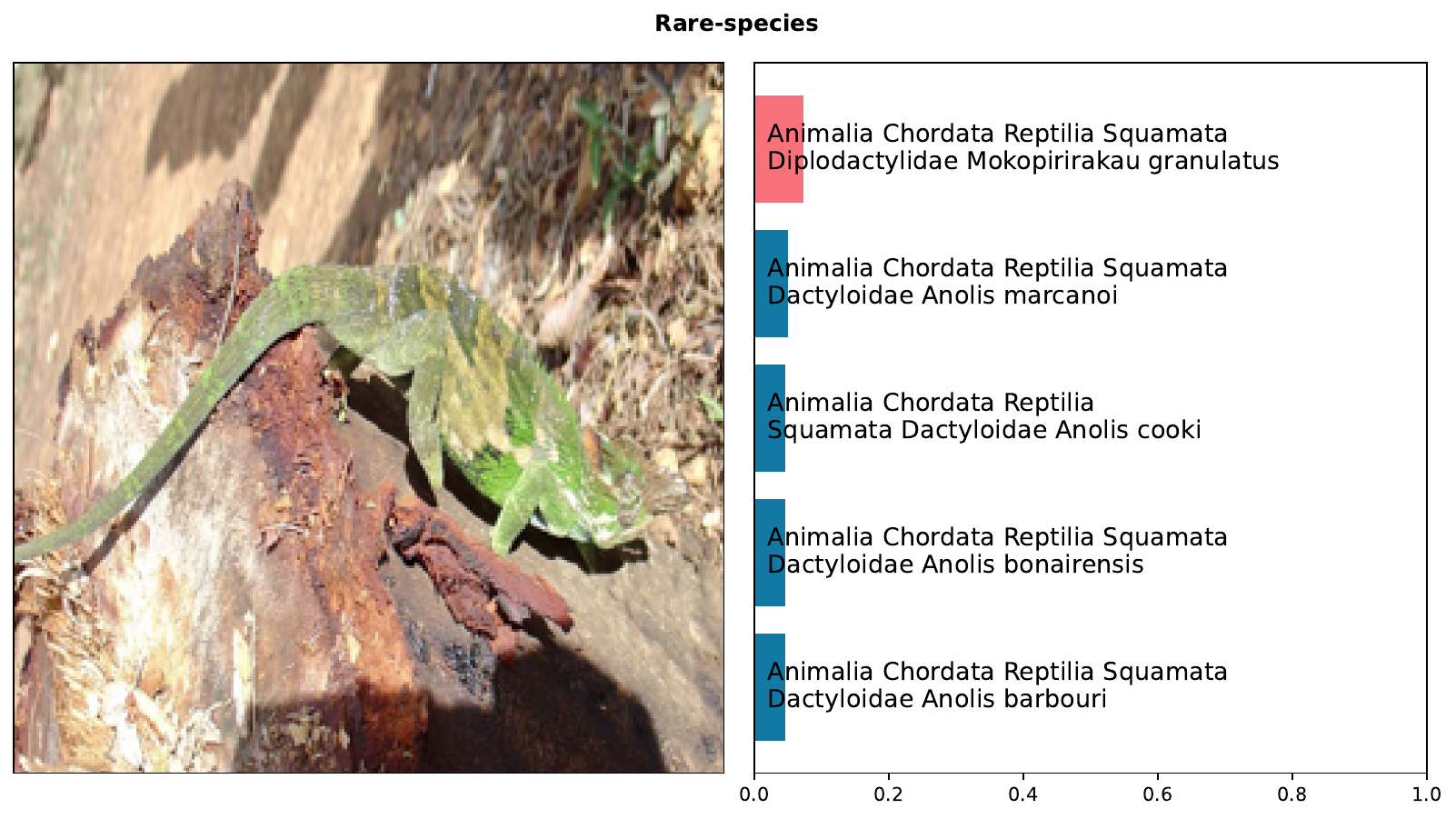}
    \end{subfigure}
    \hfill
    \begin{subfigure}[b]{0.33\textwidth}
        \includegraphics[width=\textwidth]{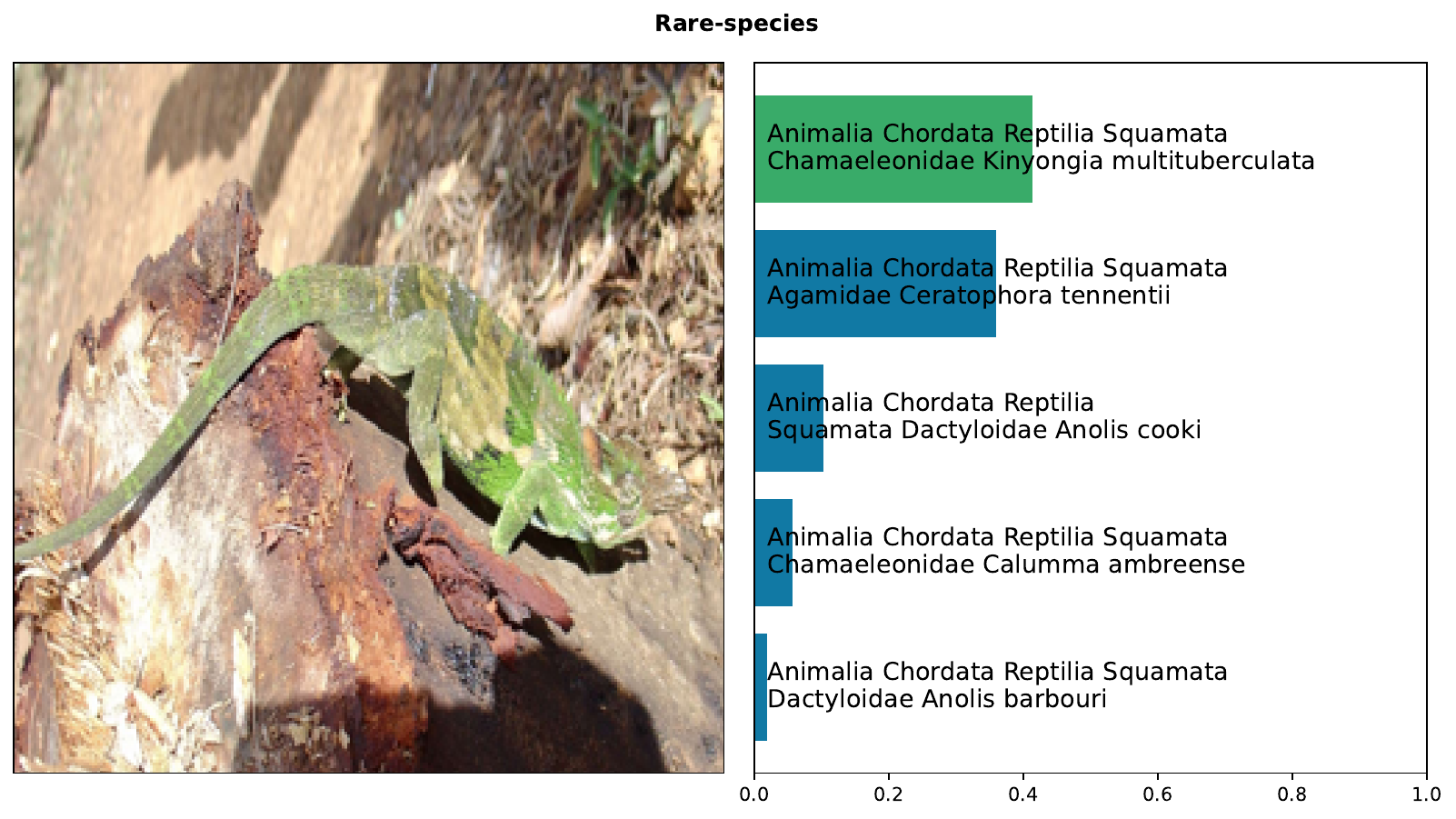}
    \end{subfigure}
    \vskip -6pt
    \caption{
    Example predictions for \modelname{} and CLIP on PlantVillage, Medicinal Leaf, PlantDoc and \rarespecies{.}
    Ground truth labels are green; incorrect predictions are red.
    Left: Correct \modelname{} predictions.
    Center, Right: Images that CLIP incorrectly labels, but \modelname{} correctly labels.
    }
    \label{fig:example-predictions2}
    \vskip -8pt
\end{figure*}

\cref{fig:example-predictions1,fig:example-predictions2} show \modelname{} and CLIP zero-shot predictions on all ten evaluation tasks.
We randomly pick examples from each dataset that \modelname{} correctly labels and example that CLIP incorrect labels but \modelname{} correctly labels.
\modelname{} performs well on a variety of tasks, including out-of-distribution images (Plankton, Medicinal Leaf) and mixes of scientific and common names (PlantVillage, PlantDoc).

\section{More Results of Text-Type}\label{app:more-text-type}

We investigated the effects of text-type during training and testing in \cref{subsec:text-type-ablation} using the \rarespecies{} task.
We present zero-shot results for all text-types on all tasks using the same procedure as in \cref{subsec:zero-shot}, where we use whatever taxonomic+common if available, otherwise whatever text-type is available.

\begin{table*}[t]
    \setlength\tabcolsep{4pt}
    \newcommand{\faded}{\color{Gray}}

    \centering
    \small
    \newcommand{\first}{\bfseries}
    \scalebox{0.93}{
    \begin{tabular}{lSSSSSSSSSSSS}
        \toprule
         & \multicolumn{4}{c}{\thead{Animals}} & \multicolumn{5}{c}{\thead{Plants \& Fungi}} & & & \\  
        \cmidrule(lr){2-5} \cmidrule(lr){6-10} 
        \thead{Training Text Type} & \vertical{Birds 525} & \vertical{Plankton} & \vertical{Insects} & \vertical{Insects 2} & \vertical{PlantNet} & \vertical{Fungi} & \vertical{PlantVillage} & \vertical{Med. Leaf} & \vertical{PlantDoc} & \vertical{Rare Species} & \multicolumn{2}{c}{Mean ($\Delta$)} \\
        \midrule
        \faded Random Guessing & \faded 0.2 & \faded 1.2 & \faded 1.0 & \faded 1.0 & \faded 4.0 & \faded 4.0 & \faded 2.6 & \faded 4.0 & \faded 3.7 & \faded 0.3 & \faded 2.2 & \\
        \midrule
        Common & 58.5 & \first 4.4 & 15.8 & 13.3 & 45.2 & 20.7 & 10.7 & 15.4 & 19.6 & 24.9 & 22.8 & -10.1\\
        Scientific & 59.7 & 3.8 & 18.7 & 11.0 & 84.8 & \first 35.3 & 12.5 & 20.3 & 13.9 & 22.3 & 28.2 & -4.7\\
        Taxonomic & 62.7 & 2.2 & 25.1 & 8.7 & 70.4 & 29.0 & 8.8 & 18.4 & 12.8 & 26.6 & 26.4 & -6.5\\
        Sci+Com & 60.2 & 2.2 & 19.2 & 12.6 & 71.5 & 24.8 & 17.6 & \first 21.5 & 20.0 & 28.0 & 27.7 & -5.2 \\
        Tax+Com & 60.2 & 2.0 & 27.4 & 11.6 & 68.4 & 19.2 & 10.4 & 19.5 & 15.8 & 30.4 & 26.4 & -6.5 \\
        Mixture & \first 65.1  & 3.5 & \first 30.6 & \first 17.3 & \first 86.3 & 32.8 & \first 19.9 & 18.7 & \first 24.5 & \first 30.9 & \first 32.9 & \text{--} \\

        \bottomrule
    \end{tabular}
    }
    \vskip -8pt
    \caption{
        Zero--shot classification top-1 accuracy for different text-types used during training.
        \textbf{Bold} indicates best accuracy.
        All models use the same architecture (ViT-B/16 vision encoders, 77-token text encoder) and are trained on the same dataset (\datasetsmall{}).
        $\Delta$ denotes the difference in mean accuracy with ``Mixture'', which is the text-type we used for \modelname{.}
    }
    \label{tab:text-type-detailed-results}
    \vskip -10pt
\end{table*}

\section{Generalized Zero-Shot Learning}\label{app:gzsl}
\citet{chao2016empirical} introduced \textit{generalized zero-shot learning}, where a model must label images of unseen classes from a set of both seen and unseen labels.
We pick out a set of 400 seen species from \datasetname{} using the same methodology as we used for the \rarespecies{} task.
We classify the same images from the \rarespecies{} task using this set of 800 labels (a mix of seen and unseen).
CLIP and OpenCLIP achieve \num{23.0}\% and \num{18.2}\% top-1 accuracy, while \modelname{} achieves \num{26.0}\% top-1 accuracy in this challenging GZSL setting.

\end{document}